\documentclass[11pt]{article}

\usepackage[final]{acl}

\usepackage{times}
\usepackage{amsmath}
\usepackage{amssymb}
\usepackage{latexsym}
\usepackage{makecell}
\usepackage{multirow}
\usepackage{booktabs}

\usepackage[T1]{fontenc}

\usepackage[utf8]{inputenc}

\usepackage{microtype}
\usepackage{inconsolata}

\usepackage{graphicx}
\usepackage{subcaption}

\usepackage{xcolor}
\usepackage{listings}

\usepackage{tikz}
\usetikzlibrary{arrows.meta,positioning,fit,shapes.multipart,calc}
\usepackage[most]{tcolorbox}

\newtcolorbox{boxB}{
    fontupper = \footnotesize,
    colframe = black,
    colback  = white,
    boxrule  = 0.8pt,
    rounded corners,
    arc = 5pt,
    left=4pt, right=4pt, top=4pt, bottom=2pt
}

\newtcolorbox{boxC}{
    fontupper = \footnotesize,
    colframe = blue!60!black,
    colback  = blue!5!white,
    boxrule  = 0.8pt,
    rounded corners,
    arc = 5pt,
    left=4pt, right=4pt, top=4pt, bottom=2pt
}

\lstdefinestyle{promptstyle}{
  basicstyle=\ttfamily\footnotesize,
  columns=fullflexible,
  breaklines=true,
  breakatwhitespace=true,
  keepspaces=true,
  showstringspaces=false,
  frame=single,
  rulecolor=\color{black!20},
  frameround=tttt,
  aboveskip=6pt,
  belowskip=0pt
}

\title{\textsc{Litmus (Re)Agent}: A Benchmark and Agentic System for Predictive Evaluation of Multilingual Models}

\author{
  Avni Mittal$^{1}$ \quad
  Shanu Kumar$^{2}$ \quad
  Sandipan Dandapat$^{3}$ \quad
  Monojit Choudhury$^{2}$ \\
  $^{1}$Microsoft Corporation, India \\
  $^{2}$Mohamed bin Zayed University of Artificial Intelligence (MBZUAI) \\
  $^{3}$Indian Institute of Technology Hyderabad \\
  \small{\texttt{avnimittal@microsoft.com} \quad
  \texttt{\{shanu.kumar, monojit.choudhury\}@mbzuai.ac.ae} \quad
  \texttt{sdandapat@cse.iith.ac.in}}
}

\begin{document}
\maketitle
\begin{abstract}
We study \emph{predictive multilingual evaluation}: estimating how well a model will perform on a task in a target language when direct benchmark results are missing. This problem is common in multilingual deployment, where evaluation coverage is sparse and published evidence is uneven across languages, tasks, and model families. We introduce a controlled benchmark of 1{,}500 questions spanning six tasks and five evidence scenarios. The benchmark separates \emph{accessible evidence} from \emph{ground truth}, enabling evaluation of systems that must infer missing results from incomplete literature evidence. We also present \textsc{Litmus (Re)Agent}, a DAG-orchestrated agentic system that decomposes queries into hypotheses, retrieves evidence, and synthesises predictions through feature-aware aggregation. Across six systems, \textsc{Litmus (Re)Agent} achieves the best overall performance, with the largest gains in transfer-heavy scenarios where direct evidence is weak or absent. These results show that structured agentic reasoning is a promising approach to multilingual performance estimation under incomplete evidence.
\end{abstract}

\begin{figure*}[h]
    \centering
    \includegraphics[width=0.8\textwidth]{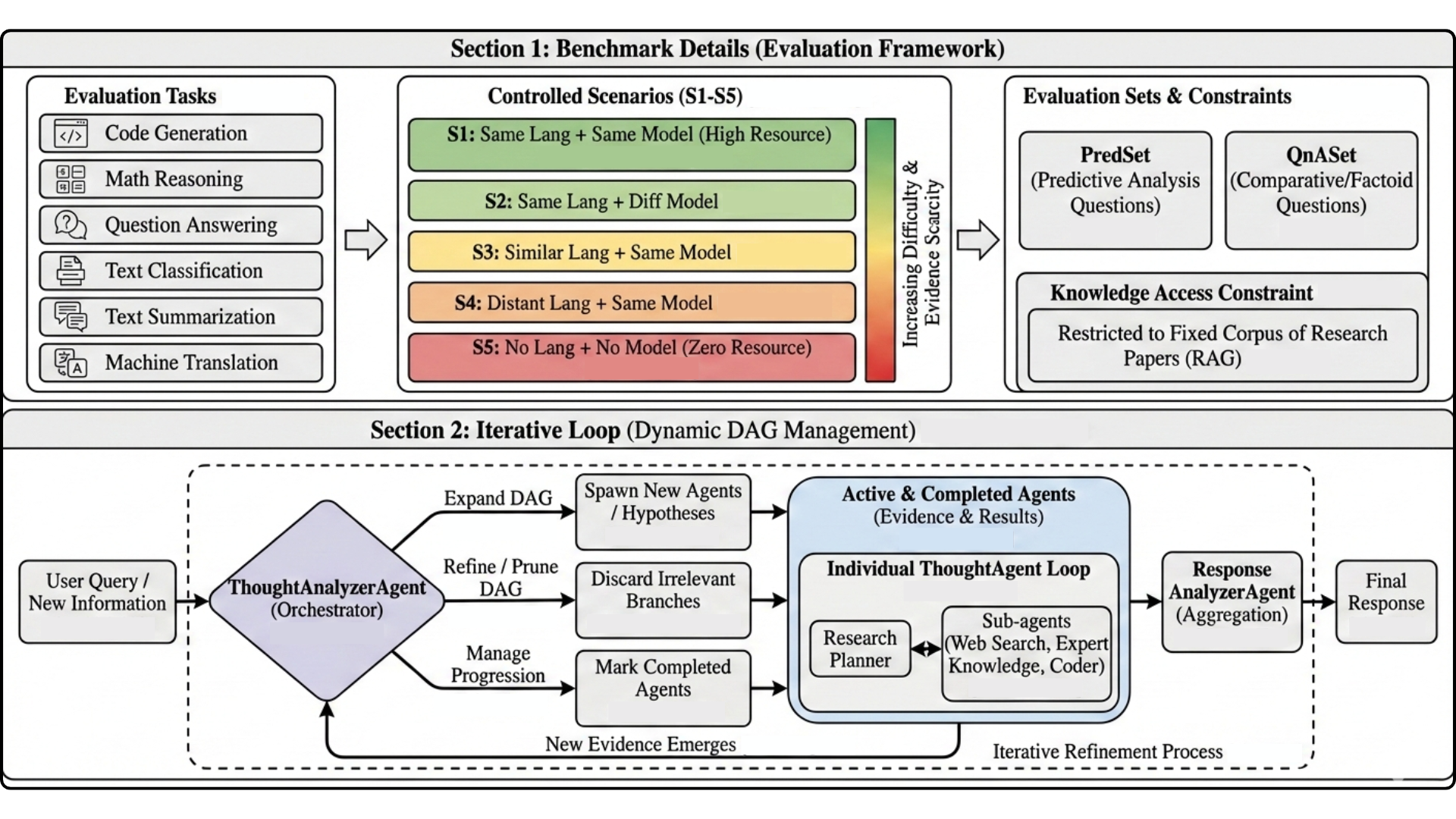}
    \caption{Overview of the benchmark and \textsc{Litmus (Re)Agent}. Top: six tasks, five controlled scenarios (S1--S5), two query types, and restricted paper-corpus access. Bottom: dynamic DAG orchestration in which specialised agents spawn and prune hypotheses, gather evidence, and aggregate results into the final response.}
    \label{fig:ui}
\end{figure*}
\section{Introduction}

Large Language Models (LLMs) are increasingly expected to operate across many tasks and many languages, yet multilingual evaluation remains highly incomplete \cite{openai2023gpt4,nllb2022}. For many task--model--language combinations, especially in low-resource settings, direct benchmark results are unavailable, scattered across papers, reported under inconsistent conditions, or too expensive to reproduce. As a result, practitioners often need to make deployment and model-selection decisions without the exact evidence they would ideally want. This creates a practical question: \emph{how will a model perform on a task in a target language when direct evaluation results are missing?} We refer to this as the \emph{Task--Model--Language} (TML) prediction problem.

Existing approaches address this problem only partially. Multilingual benchmark suites expand evaluation coverage, but still leave large portions of the TML space unobserved \cite{ruder2021xtreme,liang2022helm,nllb2022}. Predictive transfer methods use typological, representational, or information-theoretic signals \cite{lin2019choosing,nguyen2020leep,srinivasan2022litmus,tsvetkov2024information}, but usually rely on fixed features rather than reasoning over scientific evidence. LLM-as-judge methods scale evaluation, but they do not directly solve missing-result prediction from incomplete literature evidence and also raise concerns about bias and reproducibility \cite{judgebench2024}. More recent agentic systems can retrieve and synthesise information from research papers \cite{anonymous2025predictive}, but predictive reasoning under systematically restricted evidence remains underexplored. The field therefore still lacks a controlled framework for studying how systems infer missing multilingual performance from incomplete literature evidence.

To address this gap, we introduce both a benchmark and a system. First, we construct a controlled benchmark for predictive multilingual evaluation. The benchmark contains 1{,}500 questions spanning six tasks and five evidence scenarios, and separates \emph{accessible evidence} at inference time from \emph{ground truth}. Systems are evaluated using only a reduced paper corpus, while answers are defined from a larger combined corpus. This enables controlled study of predictive reasoning under varying evidence conditions. The benchmark covers two complementary capabilities: numeric performance prediction and comparative multilingual reasoning.

Second, we present \textsc{Litmus (Re)Agent}, a DAG-orchestrated agentic system for this setting. The system decomposes a query into hypotheses, retrieves citation-grounded evidence, extracts linguistic and task-level signals, and aggregates them into a final prediction. Relative to the earlier DAG-based system of \citet{anonymous2025predictive}, our version strengthens expert-knowledge retrieval, expands coder support with linguistic feature libraries, and improves prompting for more stable, expert-aligned reasoning.

We evaluate \textsc{Litmus (Re)Agent} against five baselines, including the prior DAG-based system, single-agent and non-DAG agentic variants, a direct GPT-4.1 baseline, and a general-purpose multi-agent framework. Across the full benchmark, \textsc{Litmus (Re)Agent} achieves the strongest overall performance, with the largest gains in transfer-heavy settings where direct evidence is weak or absent. We further provide analysis across tasks, evidence scenarios, metric types, and internal agent behaviours, together with a human evaluation study.

\noindent\textbf{Our contributions are threefold:} (i) we introduce a controlled benchmark for predictive multilingual evaluation under incomplete evidence, spanning six tasks, five evidence scenarios, and both numeric prediction and comparative reasoning; (ii) we present \textsc{Litmus (Re)Agent}, a DAG-orchestrated, citation-grounded system for estimating multilingual performance from incomplete literature evidence; and (iii) we provide a comprehensive empirical analysis, including comparisons against five baselines, breakdowns by task and evidence scenario, internal agent-behaviour diagnostics, and a human evaluation study.

\section{Related Work}

\subsection{Multilingual Evaluation and Benchmarks}

Multilingual evaluation has been studied through benchmark suites and task-specific datasets. Resources such as XTREME, XTREME-R, and XGLUE evaluate cross-lingual transfer across multiple languages and tasks \cite{hu2020xtreme,ruder2021xtreme,liang2020xglue}, while broader platforms such as HELM and BIG-Bench extend evaluation across diverse capabilities and models \cite{liang2022helm,srivastava2023beyond}. Task-specific benchmarks further cover program synthesis, mathematical reasoning, and summarisation \cite{chen2021evaluating,cobbe2021training,narayan2018dont}. These resources provide valuable multilingual evidence, but they do not address the problem of predicting missing results under incomplete coverage.

\subsection{Predictive Multilingual Analysis}

A separate line of work estimates performance in unseen language--task settings without running full evaluations. Early approaches used typological and corpus-based signals such as LangRank \cite{lin2019choosing}, followed by representation-based, information-theoretic, and transferability methods including Task2Vec, LEEP, and LogME \cite{achille2019task2vec,nguyen2020leep,you2021logme}. Later work introduced meta-predictors, ranking-based approaches, and multilingual deployment frameworks such as LITMUS \cite{ahuja2022beyond,srinivasan2022litmus}, while recent methods explore scalable prediction through information-theoretic modelling and language ranking \cite{tsvetkov2024information,schram2023performance,li2024languageranker}. These approaches are closely related to our setting, but they generally rely on predefined features rather than literature-grounded reasoning under controlled evidence restriction.

\subsection{Agentic and Literature-Grounded Systems}

Agentic systems augment language models with planning, tool use, and structured reasoning. ReAct and Toolformer introduced interleaved reasoning and tool invocation \cite{yao2022react,schick2023toolformer}, while graph- and multi-agent frameworks support more structured workflows \cite{wu2023gptswarm,wang2024graphagent}. Retrieval-augmented systems and scientific assistants, including PaperQA and related research agents, support citation-grounded reasoning over documents \cite{lewis2020rag,liu2023paperqa,zhang2024deepresearcher}. Most of this work focuses on literature exploration or question answering rather than multilingual performance prediction. The closest prior work is \citet{anonymous2025predictive}, which introduced a DAG-based predictive analysis system but did not provide a large, scenario-controlled benchmark for studying performance under systematically incomplete evidence.

\section{Benchmark Details}
\label{sec:benchmark}

We construct a controlled benchmark for multilingual predictive evaluation under incomplete evidence. Each instance corresponds to a Task--Model--Language query and is assigned to one of five scenarios that vary the evidence available at inference time. The key design is to separate \emph{accessible evidence} from \emph{ground truth}: systems retrieve from a reduced paper corpus, while answers are defined from a larger combined corpus.

\subsection{Benchmark Scope}

The benchmark spans six tasks: code generation, mathematical reasoning, question answering (QA/VQA), text classification (including NLI), text summarisation, and machine translation. We curate multilingual evaluation papers and extract language-to-model-family mappings, which we aggregate into a task-specific {combined mapping that represents the full answer space.
To simulate incomplete evidence, we construct a {reduced} mapping by removing a subset of papers from the combined corpus. Systems access only the reduced corpus during evaluation, while ground-truth values are taken from the combined set, enabling controlled analysis under partial evidence.
\begin{table*}[t]
\centering
\small
\setlength{\tabcolsep}{6pt}
\begin{tabular}{lccc}
\toprule
\textbf{Scenario} & \textbf{Target language} & \textbf{Target model} & \textbf{Available evidence source} \\
\midrule
S1: Same L + Same M & Observed & Observed & Direct evidence \\
S2: Same L + Diff M & Observed & Missing & Same language, other models \\
S3: Similar L + Same M & Missing & Observed & Typologically close language \\
S4: Distant L + Same M & Missing & Observed & Typologically distant language \\
S5: Diff L + Diff M & Missing & Missing & Other languages and model families \\
\bottomrule
\end{tabular}
\caption{Summary of the five benchmark scenarios. ``Observed'' means directly available in the reduced evidence corpus at inference time.}
\label{tab:scenario_summary}
\end{table*}

\subsection{Query Types: Numeric Prediction and Comparative Reasoning}

The benchmark evaluates two complementary capabilities: {numeric performance prediction} and {comparative multilingual reasoning}. \textsc{PredSet} requires predicting a scalar score for a given task, model family, and language, while \textsc{QnASet} requires comparing candidate models or languages to identify the best-performing option.
This distinction reflects practical usage: users ask both \emph{``What score should I expect?''} and \emph{``Which model is likely to be best?''}, so the benchmark covers both regression and comparative reasoning.

\begin{figure}[ht]
\centering
\resizebox{0.47\textwidth}{!}{%
\begin{boxB}
{\footnotesize \textbf{Numeric Prediction (Code Generation):}}\\
{\footnotesize \textbf{Q1:} \textit{What is the performance of DeepSeek-Coder-V2 on Code Generation for Nepali?}}\\
{\footnotesize \textbf{Answer:} \textcolor{red}{1.5 (pass@1)}}\\[0.12cm]
{\footnotesize \textbf{Numeric Prediction (Classification/NLI):}}\\
{\footnotesize \textbf{Q2:} \textit{What score does GPT-4-Turbo achieve for Classification NLI in Oromo?}}\\
{\footnotesize \textbf{Answer:} \textcolor{red}{10.7}}\\[0.12cm]
{\footnotesize \textbf{Numeric Prediction (Machine Translation):}}\\
{\footnotesize \textbf{Q3:} \textit{How well does IndicTrans2 perform on Machine Translation for Sanskrit?}}\\
{\footnotesize \textbf{Answer:} \textcolor{red}{51.6 (chrF++)}}\\[0.12cm]
{\footnotesize \textbf{Comparative Reasoning (Text Summarisation):}}\\
{\footnotesize \textbf{Q4:} \textit{Which model achieves the highest score for Text Summarization in Kirundi?}}\\
{\footnotesize \textbf{Answer:} \textcolor{red}{Qwen2-7B}}\\[0.12cm]
{\footnotesize \textbf{Comparative Reasoning (QA/VQA):}}\\
{\footnotesize \textbf{Q5:} \textit{What is the top-performing model for QA VQA in Tigrinya?}}\\
{\footnotesize \textbf{Answer:} \textcolor{red}{TiRoBERTa}}
\end{boxB}
}
\caption{Illustrative questions from the two benchmark subsets: \textsc{PredSet} for numeric prediction and \textsc{QnASet} for comparative reasoning.}
\label{box:examples}
\end{figure}

\subsection{Scenario Design}

Table~\ref{tab:scenario_summary} summarises the five evidence scenarios used in the benchmark. Throughout, \textbf{same model} refers to the same \textbf{model family}, rather than the exact checkpoint identifier. \textbf{S1} is the strongest-evidence setting, where both the target language and target model family are observed in the reduced corpus, allowing direct retrieval for the exact Task--Model--Language combination. \textbf{S2} keeps the target language observed but removes the target model family, so prediction must transfer across models within the same language. \textbf{S3} and \textbf{S4} both remove the target language while retaining evidence for the same model family: \textbf{S3} allows transfer from a typologically close language, whereas \textbf{S4} allows only distant-language evidence. \textbf{S5} is the weakest-evidence setting, where both the target language and target model family are absent, requiring transfer across both languages and model families.








\subsection{Question Construction}

Questions are generated for each task--scenario block from language-to-model-family mappings. Prompts are constrained to use only languages and model families present in the mapping, and exclude programming languages for code generation. Each block includes both numeric-prediction and comparative-reasoning questions. The full prompt template is provided in Appendix~\ref{appendix:benchmark-creation} (Figure~\ref{fig:prompt_template}).

\subsection{Ground Truth and Metric Normalisation}
\label{sec:gt_norm}

Ground-truth values are automatically extracted from the paper collection and manually validated. We record the associated metric for each question (e.g., \textit{pass@1}, Accuracy, F1, BLEU) and normalise all scores to a common 0--100 scale. Values in $[0,1]$ are multiplied by 100, while other metrics undergo task-specific linear scaling. Evaluation is restricted to compatible metric families within each task to ensure valid comparisons.

\subsection{Language Similarity for Transfer Scenarios}

For Scenarios~3 and~4, language similarity is defined using cosine distance over \texttt{lang2vec} features, with language pairs split into \emph{close} and \emph{distant} groups via a percentile-based threshold. Full details are provided in Appendix~\ref{appendix:language-similarity}.

\subsection{Dataset Statistics}
\label{sec:benchmark_stats}

The benchmark contains 1{,}500 questions across six tasks and five scenarios, with 50 questions per task--scenario block (25 \textsc{PredSet} and 25 \textsc{QnASet}). The reduced evidence set is constructed to ensure all scenarios remain instantiable for every task.
Figure~\ref{fig:combined_reduced_stats} compares the combined and reduced paper collections. The reduced set has lower language and model-family coverage, creating controlled evidence scarcity while retaining ground truth in the combined set. Additional statistics are provided in Appendix~\ref{appendix:benchmark-creation}.

\begin{figure}[t]
    \centering
    \includegraphics[width=0.5\textwidth]{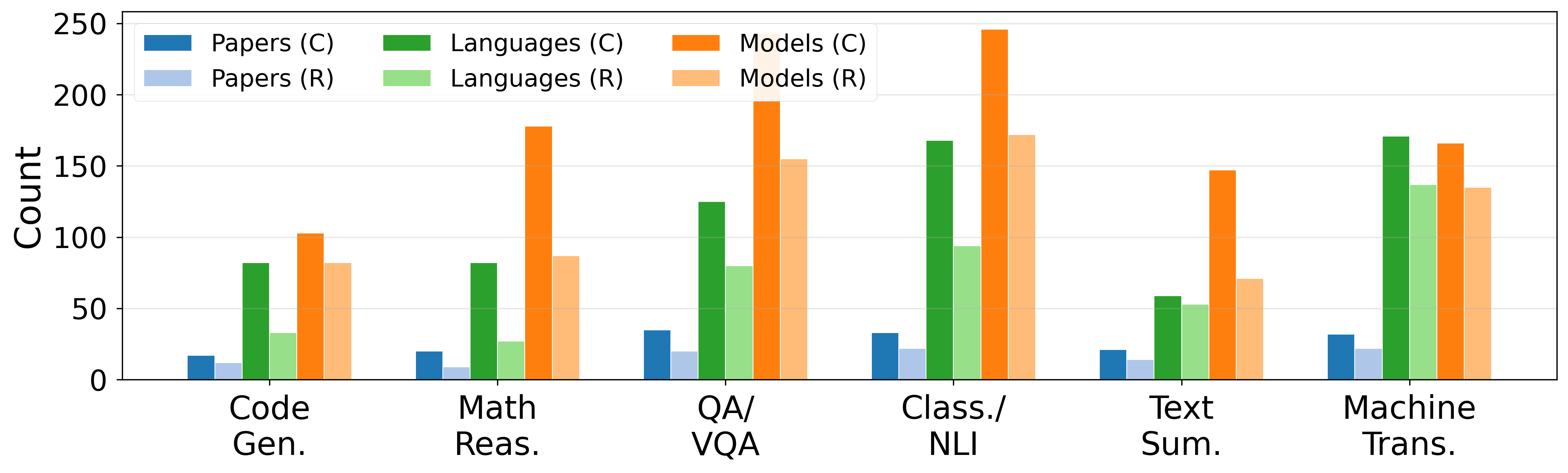}
    \caption{Combined versus reduced corpus statistics per task.}
    \label{fig:combined_reduced_stats}
\end{figure}

\section{\textsc{Litmus (Re)Agent}}
\label{sec:system}

We build on a DAG-based multi-agent architecture originally proposed for multilingual performance prediction \cite{anonymous2025predictive}. That system showed that decomposing evaluation queries into parallel hypothesis threads handled by specialised \emph{Thought Agents} yields more grounded predictions than monolithic prompting. We retain the core DAG structure but introduce targeted improvements to expert-knowledge retrieval, code execution with linguistic features, and prompt design. These modifications produce \textsc{Litmus (Re)Agent}, enabling reliable predictive analysis under restricted evidence by aligning system behaviour with expert multilingual evaluation workflows. Implementation details are provided in Appendix~\ref{appendix:implementation}.

\subsection{Base Architecture}
\label{sec:base_arch}

The system follows a DAG-based orchestration pattern \cite{zhang2024diagram}. A natural-language query (e.g., \textit{``How will GPT-4o perform on machine translation in Nepali?''}) is processed by a \textit{MainAgent}, which determines whether it is new or a follow-up. For new queries, a \textit{ThoughtCreatorAgent} decomposes the problem into hypotheses and spawns one \textit{ThoughtAgent} per hypothesis as a node in the DAG.
Each ThoughtAgent operates as a coordinated group of specialised sub-agents: 
\textbf{Research Planner} (plans evidence acquisition and coordinates analysis), 
\textbf{Web Search and Crawl} (retrieves and parses documents), 
\textbf{Expert Knowledge} (queries a curated multilingual knowledge base), 
\textbf{Coder} (writes and executes analysis scripts, including regression models), and 
\textbf{Reporter} (summarises the hypothesis outcome).

A \textit{ThoughtAnalyzerAgent} monitors active hypotheses, spawning new ones when gaps appear and pruning redundant branches. Once all hypotheses are resolved, a \textit{ResponseAnalyzerAgent} aggregates the results into a final response containing the predicted outcome, supporting citations, and a structured rationale. This modular design limits context-window saturation while preserving interpretable reasoning, as each hypothesis maintains an independent evidence trail.
\subsection{Improvements in \textsc{Litmus (Re)Agent}}
\label{sec:improvements}

Relative to the base system, we introduce three improvements that enhance planning, retrieval, and validation in line with expert workflows.

\paragraph{Enriched Expert Knowledge.}
The base system relied on a small set of task-specific papers. We expand this to a broader corpus of multi-task cross-lingual evaluation studies, reflecting the transferability of predictive signals \cite{ahuja-etal-2022-multi, srinivasan2022litmus}. The knowledge base encodes multilingual evaluation strategies, including language--task--model observations, failure modes, and expert analysis patterns, enabling more informed prediction planning.

\paragraph{Enhanced Code Execution with Linguistic Feature Libraries.}
The Coder agent previously had limited library access and lacked structured linguistic resources. We extend it with \texttt{lang2vec} and the URIEL typological database \cite{littell2017uriel}, providing syntactic, phonological, geographic, and family-level features for over 7{,}000 languages. These enable language distance computation, feature extraction, and feature-informed regression models used in multilingual research.

\paragraph{Prompt Stabilisation and Expert-Aligned Reasoning.}
Prompts are redesigned to reinforce the analysis process and effective use of system capabilities. Instructions guide agents to follow expert-style workflows, including hypothesis generation, evidence retrieval, and feature-based analysis, while requiring cited evidence and careful tool usage. This reduces failure modes such as invalid tool calls, brittle code execution, and hallucinated citations.

\subsection{Evaluation Setup}

We evaluate all methods on the benchmark described in Section~\ref{sec:benchmark}. During evaluation, retrieval is restricted to the \emph{reduced} paper corpus, while ground truth is computed from the \emph{combined} corpus (Section~\ref{sec:gt_norm}). This restriction is part of the evaluation protocol rather than a limitation of the system itself: it prevents direct access to held-out results and enables controlled study of predictive reasoning under incomplete evidence.

We compare \textsc{Litmus (Re)Agent} against five baselines: (i) \textbf{LITMUS++}, the prior DAG-based system without the improvements introduced in this work; (ii) \textbf{ThoughtAgent}, a single multi-agent group chat without DAG decomposition; (iii) \textbf{Single Agent}, a ReAct-style agent with direct tool access but no multi-agent coordination; (iv) a \textbf{GPT-4.1 direct baseline} with no agentic scaffolding; and (v) \textbf{Magentic-One} \cite{fourney2024magentic}, a general-purpose multi-agent framework. All agentic systems use GPT-4.1 as the underlying backbone. Additional implementation details are provided in Appendix~\ref{appendix:systems-compared}.


\begin{table*}[t]
\centering
\small
\setlength{\tabcolsep}{7pt}
\begin{tabular}{l cccccc | cccccc}
\hline
 & \multicolumn{6}{c|}{\textbf{PredSet MAE} ($\downarrow$)} & \multicolumn{6}{c}{\textbf{QnASet Accuracy (\%)} ($\uparrow$)} \\
\textbf{Task} & \textbf{LRA} & \textbf{L++} & \textbf{TA} & \textbf{SA} & \textbf{GPT} & \textbf{M1} & \textbf{LRA} & \textbf{L++} & \textbf{TA} & \textbf{SA} & \textbf{GPT} & \textbf{M1} \\
\hline
Code Gen.    & \textbf{14.2} & 21.1 & 18.9 & 17.8 & 39.8 & 27.0 & \textbf{55.3} & 25.4 & 39.5 & 51.8 & 27.2 & 41.2 \\
Math Reas.   & 20.0 & 28.8 & 27.5 & 29.6 & \textbf{18.2} & 26.8 & 15.4 & 10.6 & 13.0 & 9.8 & \textbf{18.7} & 7.3 \\
QA/VQA       & 9.5  & \textbf{8.7} & 14.7 & 12.4 & 9.6  & 14.1 & 8.0 & \textbf{22.4} & 8.0 & 16.8 & 16.0 & 5.6 \\
Class./NLI   & \textbf{7.9} & 9.6 & 11.6 & 11.9 & 9.9  & 12.0 & \textbf{16.0} & 7.2 & 6.4 & 13.6 & 9.6 & 2.4 \\
Text Sum.    & 5.1  & 6.9 & 7.0  & \textbf{4.7} & 6.8  & 8.3 & 21.0 & 13.4 & 10.9 & \textbf{21.8} & 15.1 & 1.7 \\
Machine Tr.  & \textbf{6.2} & 6.4 & 9.9 & 7.5 & 8.3 & 10.0 & 16.8 & 12.0 & \textbf{19.2} & 9.6 & 9.6 & 4.0 \\
\hline
\textit{Overall} & \textbf{10.4} & 12.7 & 15.1 & 14.5 & 16.5 & 17.2 & \textbf{21.6} & 15.1 & 15.9 & 20.1 & 15.9 & 10.0 \\
\hline
\end{tabular}
\caption{Task-level results on the benchmark. Left: PredSet mean absolute error (MAE; lower is better). Right: QnASet accuracy (higher is better). LRA = Litmus (Re)Agent, L++ = LITMUS++, TA = ThoughtAgent, SA = Single Agent, GPT = GPT-4.1, and M1 = Magentic-One.}
\label{tab:task_results_combined}
\end{table*}

\section{Results}
\label{sec:results}

We evaluate \textsc{Litmus (Re)Agent} on the full 1{,}500-question benchmark and compare it against five baselines: LITMUS++, ThoughtAgent, Single Agent, GPT-4.1 direct, and Magentic-One. All agentic systems use GPT-4.1 as the backbone LLM.

\subsection{Overall Performance}

Tables~\ref{tab:task_results_combined} summarise the main findings. On \textsc{PredSet}, \textsc{Litmus (Re)Agent} achieves an overall MAE of \textbf{10.3}, followed by LITMUS++ (12.7), Single Agent (14.5), ThoughtAgent (15.1), GPT-4.1 (16.4), and Magentic-One (17.2). On \textsc{QnASet}, \textsc{Litmus (Re)Agent} reaches \textbf{21.6\%} accuracy, compared to 18.9\% for the Single Agent, 15.9\% for both ThoughtAgent and GPT-4.1, 15.1\% for LITMUS++, and 10.0\% for Magentic-One. These results show that \textsc{Litmus (Re)Agent} improves both numeric prediction and comparative reasoning over all baselines, and that the three targeted improvements introduced in this work account for the gain over LITMUS++ (2.4 MAE reduction). Notably, the Single Agent (14.5) outperforms the ThoughtAgent (15.1) on MAE, suggesting that the multi-agent group chat overhead does not always help without DAG-level decomposition.

\subsection{Per-Task Analysis}

Table~\ref{tab:task_results_combined} present the per-task results. On \textsc{PredSet}, \textsc{Litmus (Re)Agent} achieves the lowest MAE on four of six tasks, with the largest gain in code generation (14.2 vs.\ 39.8 for GPT-4.1), where evidence decomposition and feature-informed analysis are particularly effective. Both the ThoughtAgent (18.9) and Single Agent (17.8) substantially outperform GPT-4.1 on this task, indicating that tool use alone provides benefits even without DAG structure. Text summarisation is relatively easy for all systems (MAE 4.1--7.8), with the Single Agent achieving the best result (4.1). Mathematical reasoning remains challenging, and is the only task where GPT-4.1 performs best (18.2), likely reflecting weaker dependence on external evidence. Magentic-One consistently performs poorly across tasks, suggesting that general-purpose orchestration does not transfer well without task-specific design.

On \textsc{QnASet}, \textsc{Litmus (Re)Agent} achieves the highest overall accuracy (21.6\%) and leads on code generation (55.3\%) and classification/NLI (16.0\%). Other systems achieve the best performance on individual tasks (e.g., ThoughtAgent on machine translation, LITMUS++ on QA/VQA, Single Agent on summarisation), but overall accuracy remains low across tasks. This indicates that comparative multilingual reasoning under incomplete evidence is substantially harder than numeric prediction. A full task--scenario breakdown is provided in Appendix Table~\ref{tab:task_scenario_mae}.

\begin{table*}[t]
\centering
\small
\setlength{\tabcolsep}{6pt}
\begin{tabular}{l cccccc | cccccc}
\hline
 & \multicolumn{6}{c|}{\textbf{MAE} ($\downarrow$)} & \multicolumn{6}{c}{\textbf{Accuracy \%} ($\uparrow$)} \\
\textbf{Scenario} & \textbf{LRA} & \textbf{L++} & \textbf{TA} & \textbf{SA} & \textbf{GPT} & \textbf{M1} & \textbf{LRA} & \textbf{L++} & \textbf{TA} & \textbf{SA} & \textbf{GPT} & \textbf{M1} \\
\hline
S1: Same L+M       & \textbf{9.4}  & 13.5 & 17.9 & 14.6 & 14.5 & 15.1 & \textbf{29.3} & 17.3 & 26.7 & 22.7 & 12.0 & 22.7 \\
S2: Same L, Diff M & \textbf{10.3} & 12.8 & 15.7 & 15.7 & 16.0 & 16.9 & 14.7 & 12.0 & 9.3 & \textbf{22.0} & 12.7 & 6.7  \\
S3: Similar L+M    & \textbf{9.0}  & 10.3 & 11.2 & 12.5 & 15.4 & 14.7 & \textbf{33.3} & 21.1 & 19.7 & 27.9 & 29.9 & 14.3 \\
S4: Distant L+M    & \textbf{11.7} & 12.2 & 16.3 & 15.0 & 19.1 & 23.5 & \textbf{13.8} & 13.1 & 11.7 & 13.8 & 11.0 & 2.1  \\
S5: Diff L+M       & \textbf{11.9} & 15.7 & 14.7 & 15.6 & 17.7 & 16.7 & \textbf{16.6} & 11.5 & 11.5 & 13.7 & 13.7 & 3.6  \\
\hline
\textit{Overall}   & \textbf{10.4} & 12.7 & 15.1 & 14.5 & 16.5 & 17.2 & \textbf{21.6} & 15.1 & 15.9 & 20.1 & 15.9 & 10.0 \\
\hline
\end{tabular}
\caption{PredSet MAE and QnASet accuracy by scenario. Abbreviations as in Table~\ref{tab:task_results_combined}. Best per row in \textbf{bold}.}
\label{tab:mae_per_scenario}
\end{table*}

\subsection{Per-Scenario Analysis}

Table~\ref{tab:mae_per_scenario} show results across the five evidence scenarios. \textsc{Litmus (Re)Agent} achieves the lowest MAE in all scenarios. The largest improvement appears in S1, where direct evidence must be correctly retrieved and synthesised (9.4 vs.\ 17.9 for ThoughtAgent). In transfer-heavy settings, performance differences widen: in S4, \textsc{Litmus (Re)Agent} maintains a low MAE (11.7) while Magentic-One degrades sharply (23.5), and the ThoughtAgent and Single Agent perform similarly (16.3 vs.\ 15.0), indicating limited benefit from multi-agent group chat without DAG decomposition.

On \textsc{QnASet}, \textsc{Litmus (Re)Agent} achieves the highest overall accuracy (21.6\%) and leads in four of five scenarios. Gains are most pronounced in S3 (33.3\%), suggesting that hypothesis decomposition is particularly effective for transfer from related languages. In S2, the Single Agent achieves the highest accuracy (22.0\%), while the ThoughtAgent lags (9.3\%), indicating that group chat overhead can be counterproductive for simpler cross-model transfer. Magentic-One shows a sharp drop in S4 and S5 (2.1\% and 3.6\%), highlighting the importance of domain-specific design in low-evidence settings.
As in the per-task results, \textsc{PredSet} MAE is lowest in S3 rather than S1, indicating that the scenarios do not form a strict difficulty ordering. Instead, they represent different evidence configurations, where close-language transfer can provide cleaner signals than noisy direct evidence.


\begin{table}[t]
\centering
\begin{tabular}{l r}
\hline
\textbf{Metric} & \textbf{Value} \\
\hline
Thought faithfulness & 93.9\% \\
Capability compliance & 70.0\% \\
Web search relevance & 83.5\% \\
Feature correctness & 91.6\% \\
Code execution success & 30.6\% \\
\hline
\end{tabular}
\caption{Agent reasoning quality metrics for \textsc{Litmus (Re)Agent}, computed over 1{,}730 conversations and 5{,}184 generated hypotheses.}
\label{tab:agent_quality}
\vspace{-0.5em}
\end{table}

\subsection{Agent Reasoning Quality}
\label{sec:agent_quality}

We analyse the internal behaviour of \textsc{Litmus (Re)Agent} using five diagnostic metrics over 1{,}730 conversations (Table~\ref{tab:agent_quality}): hypothesis alignment with expert guidance (\textit{thought faithfulness}), adherence to tool capabilities (\textit{capability compliance}), relevance of retrieved documents (\textit{web search relevance}), validity of selected features (\textit{feature correctness}), and successful code execution. The system performs strongly on thought faithfulness, web search relevance, and feature correctness, but more weakly on code execution (30.6\%). This suggests that hypothesis generation and evidence grounding are generally reliable, while program execution remains a bottleneck. Allowing retries raises execution success to 54.6\%, but substantial room for improvement remains.

Appendix Figure~\ref{fig:algo_heatmap} shows that the Coder agent primarily uses Random Forest, Ridge Regression, and XGBoost across tasks, while Figure~\ref{fig:feature_heatmap} highlights extensive use of typological and phylogenetic features from \texttt{lang2vec} alongside preprocessing and model-based signals. Additional breakdowns and scenario-wise code quality metrics are reported in Appendix~\ref{appendix:benchmark-creation}.

\begin{figure}[t]
    \centering
    \includegraphics[width=0.48\textwidth]{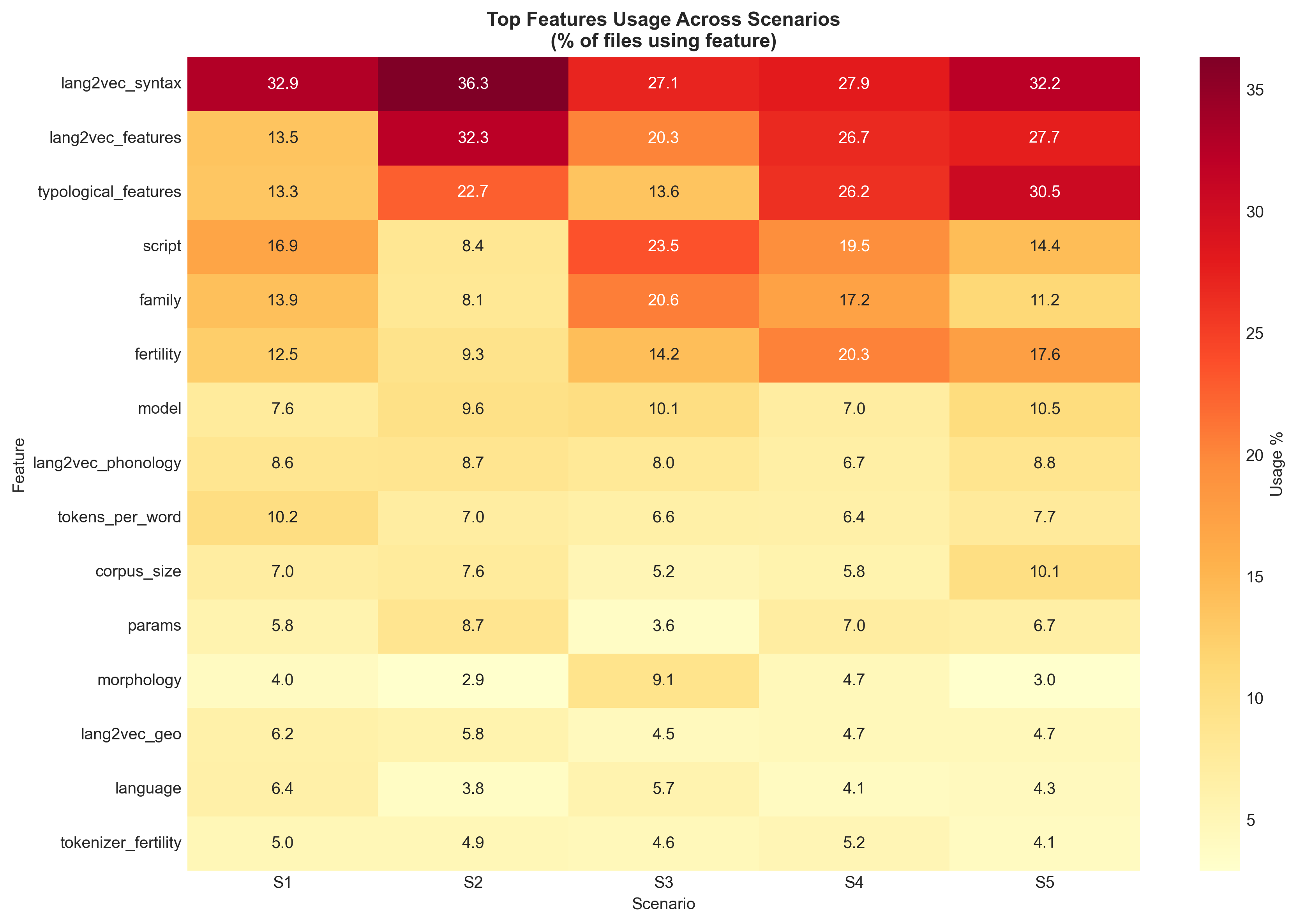}
    \caption{Feature usage heatmap across tasks. Rows represent feature categories (typological, preprocessing, model-based); columns represent tasks. Darker cells indicate higher usage.}
    \label{fig:feature_heatmap}
\end{figure}

We further assess output quality using an LLM-judge framework that scores each system's responses on four dimensions (1--5 scale): predictive plausibility, feature selection, coherence, and citation emphasis. Figure~\ref{fig:text_quality_radar} shows the overall averages across all tasks. \textsc{Litmus (Re)Agent} achieves the highest scores on predictive plausibility (4.6) and feature selection (4.8), reflecting effective hypothesis formation and linguistic feature usage. GPT-4.1 scores highest on coherence (5.0) and citation emphasis (4.3), likely because its single-pass output is more consistently formatted. The ThoughtAgent closely tracks \textsc{Litmus (Re)Agent} on plausibility and feature selection, while LITMUS++ shows markedly lower scores across all dimensions, confirming the impact of the three proposed improvements. A detailed per-task breakdown is provided in Appendix Table~\ref{tab:mae_per_scenario}.

\begin{figure}[t]
    \centering
    \includegraphics[width=0.48\textwidth]{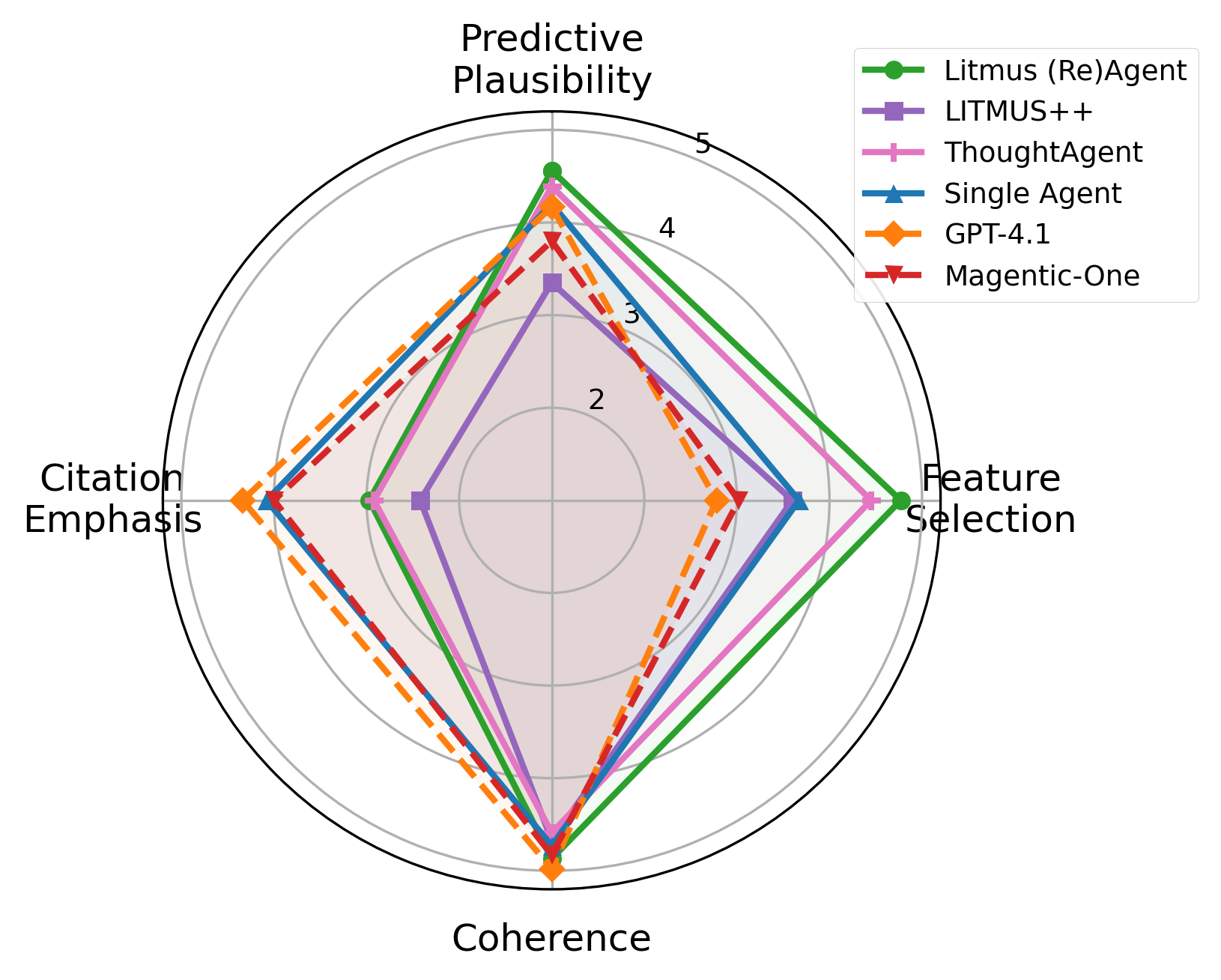}
    \caption{LLM-judge quality metrics (1--5) averaged across all tasks for each system. \textsc{Litmus (Re)Agent} leads on predictive plausibility and feature selection; GPT-4.1 leads on coherence and citation emphasis.}
    \label{fig:text_quality_radar}
\end{figure}

\subsection{Ablation Study: Backbone LLM}
\label{sec:ablation}

We evaluate the impact of the backbone LLM to assess whether gains arise from the DAG-based architecture or the underlying model. We replace GPT-4.1 with DeepSeek-V3, Llama-3.3-70B-Instruct, and o4-mini, and evaluate on code generation across S1, S3, and S5.

Table~\ref{tab:ablation_backbone} reports LLM-judge metrics (1--5 scale) on shared queries. GPT-4.1 achieves the best overall performance, particularly on predictive plausibility and feature selection. o4-mini is closest, matching coherence but trailing on citation emphasis. DeepSeek-V3 performs well in S3 but drops in S1 and S5, while Llama-3.3-70B performs worst, especially on citation grounding.
Overall, stronger backbones improve reasoning quality, but consistent trends across models indicate that gains primarily arise from the structured, evidence-grounded workflow rather than the backbone alone.

\begin{table}[t]
\centering
\small
\setlength{\tabcolsep}{2.5pt}
\begin{tabular}{l c cccc}
\hline
\textbf{Backbone} & \textbf{Scen.} & \textbf{Pred.} & \textbf{Feat.} & \textbf{Coh.} & \textbf{Cite.} \\
\hline
\multirow{3}{*}{GPT-4.1}
 & S1 & \textbf{4.8} & \textbf{4.9} & 4.9 & \textbf{3.2} \\
 & S3 & \textbf{4.7} & \textbf{4.7} & 4.9 & \textbf{2.9} \\
 & S5 & 4.5 & \textbf{4.9} & \textbf{5.0} & \textbf{3.0} \\
\hline
\multirow{3}{*}{o4-mini}
 & S1 & 4.0 & 4.4 & \textbf{5.0} & 2.8 \\
 & S3 & 4.2 & \textbf{4.7} & 4.8 & 2.7 \\
 & S5 & \textbf{4.4} & 4.8 & \textbf{5.0} & 2.6 \\
\hline
\multirow{3}{*}{DeepSeek-V3}
 & S1 & 3.6 & 3.7 & 4.6 & 2.0 \\
 & S3 & 4.3 & 4.3 & 4.9 & 2.3 \\
 & S5 & 4.0 & 3.8 & 4.9 & 2.4 \\
\hline
\multirow{3}{*}{Llama-3.3-70B}
 & S1 & 3.4 & 3.8 & 4.2 & 1.1 \\
 & S3 & 3.6 & 3.3 & 4.1 & 1.1 \\
 & S5 & 3.9 & 3.6 & 4.4 & 1.3 \\
\hline
\end{tabular}
\caption{Backbone LLM ablation on code generation (S1, S3, S5). Quality metrics from LLM-judge evaluation (1--5 scale) computed on common questions across all models. Pred.\ = Predictive Plausibility, Feat.\ = Feature Selection, Coh.\ = Coherence, Cite.\ = Citation Emphasis.}
\label{tab:ablation_backbone}
\end{table}

\subsection{Human Evaluation Study}
\label{sec:human_eval}

We conduct a within-subjects study with 8 participants (4 novice, 4 expert), each answering 10 code-generation prediction questions: 5 using general LLM tools (ChatGPT, Gemini, Claude) and 5 using \textsc{Litmus (Re)Agent}, under the same constrained paper corpus. Experts have 1--5+ years of NLP experience; novices are undergraduate students. For each question, participants report predictions, reasoning, confidence, and rate five Likert dimensions: confidence (0--5), and interpretability, explainability, justification, and actionability (1--5). Time per question is recorded.

Table~\ref{tab:human_eval} summarises results. Participants consistently rate predictions higher with \textsc{Litmus (Re)Agent}, with the largest gains in justification (+1.0) and actionability (+0.9), reflecting improved reasoning and evidence grounding. Novices benefit more overall (e.g., confidence $+1.0$, actionability $+1.0$), while experts show the largest gain in justification (+1.1). Time increases for novices (13.7 vs.\ 5.9 min) but remains stable for experts (10.3 vs.\ 10.7 min). In a final survey, 5 of 7 participants report improved prediction accuracy. Figure~\ref{fig:human_eval_radar} shows quality profiles by participant type; full results are in Appendix~\ref{appendix:human-eval-details}.

\begin{figure}[t]
    \centering
    \includegraphics[width=0.48\textwidth]{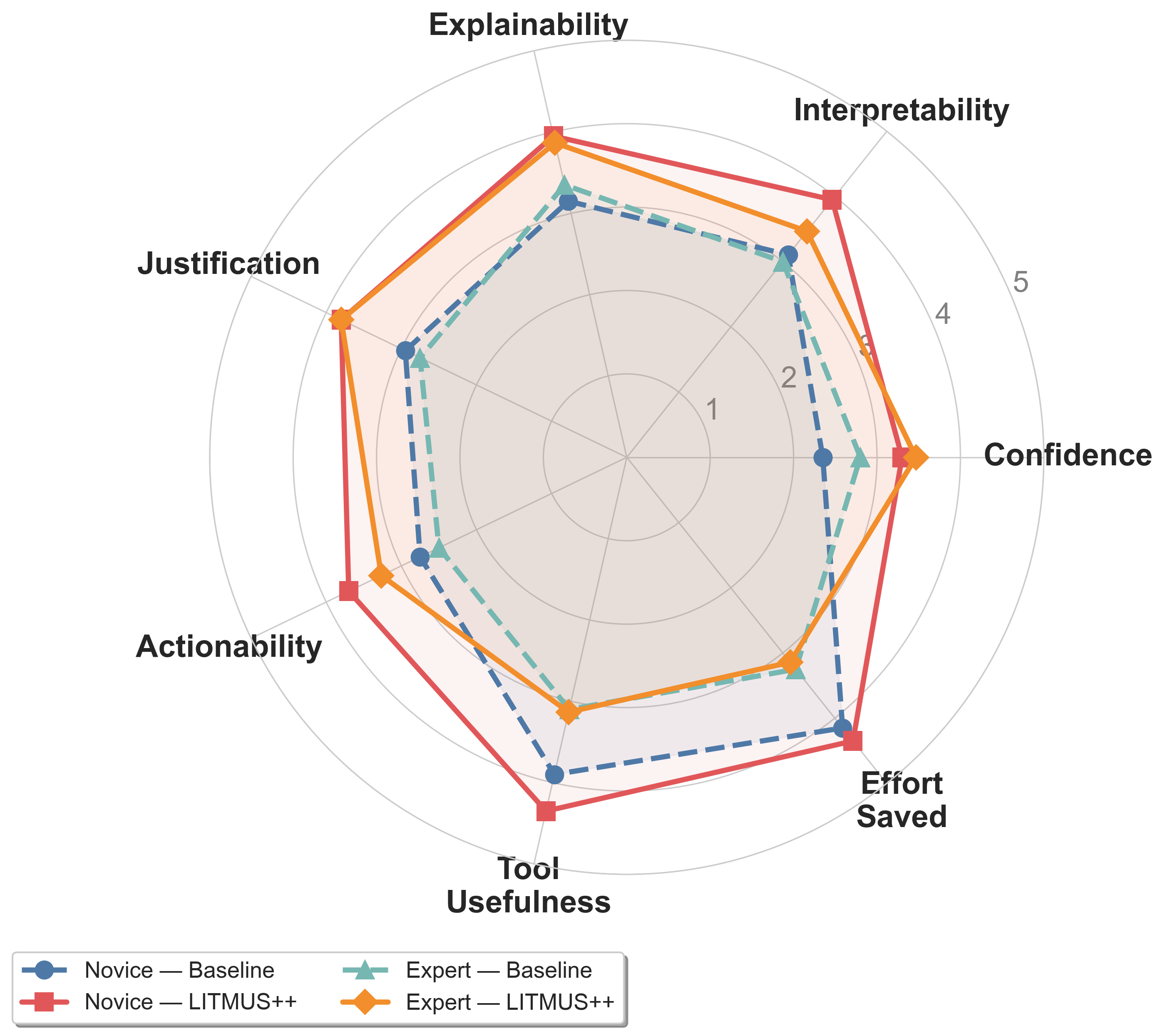}
    \caption{Human evaluation quality metrics by participant type (novice vs.\ expert) and phase (baseline vs.\ \textsc{Litmus (Re)Agent}).}
    \label{fig:human_eval_radar}
\end{figure}

\section{Conclusion}
\label{sec:conclusion}

We presented \textsc{Litmus (Re)Agent} and a controlled benchmark for multilingual performance prediction under incomplete evidence, spanning six tasks and diverse evidence scenarios. \textsc{Litmus (Re)Agent} combines DAG-based agent orchestration with citation-grounded retrieval, linguistic feature modelling, and regression-based validation to enable structured predictive analysis.
Our results show that structured, hypothesis-driven workflows consistently improve performance, particularly in transfer-heavy settings where direct evidence is limited. We further find that multi-agent group chat alone is insufficient without DAG-level decomposition, highlighting the importance of explicit reasoning structure.
Overall, this work demonstrates that agentic systems can support reliable and transparent multilingual performance estimation under realistic evidence constraints, and provides a reusable benchmark for future research in predictive evaluation.

\section{Limitations}
\label{sec:limitations}

This work has three main limitations. \textbf{System:} all agentic experiments use GPT-4.1, so generalisation to smaller or open-source backbones is unverified; Coder-agent execution success is moderate (30.6\% at the conversation level), and end-to-end latency remains higher than that of direct prompting. \textbf{Benchmark:} coverage spans six tasks but not multimodal reasoning, dialogue, or safety; predictions depend on a fixed, curated corpus; and paper-reported ground truth may include measurement noise and cross-study inconsistency. \textbf{Comparison:} while the six-system evaluation provides a broad comparison, LITMUS++, Magentic-One, and the ThoughtAgent use different internal coordination strategies, so gains over them reflect both architectural and prompt-level differences that are not fully disentangled. 

\section*{Ethical Considerations}

\textsc{Litmus (Re)Agent} is intended to support multilingual evaluation planning, not to replace direct benchmarking when high-stakes decisions require ground-truth measurement. Predictive estimates can be useful for prioritisation and early-stage analysis, but they may inherit biases from the published literature, including uneven language coverage, inconsistent reporting standards, and systematic underrepresentation of low-resource settings. To mitigate these risks, we evaluate under controlled evidence restriction, retain explicit citations to accessible evidence, and emphasise that predictions should be treated as decision-support signals rather than authoritative substitutes for benchmark results.

\bibliography{custom}
\appendix

\section{Agent Lifecycle Details}
\label{appendix:agent-lifecycle}

Each \textit{ThoughtAgent} transitions between three core states:
\textbf{Active}: investigating a hypothesis with assigned tools;
\textbf{Completed}: finished investigation and returned validated evidence;
\textbf{Discarded}: pruned when deemed irrelevant, redundant, or divergent.
State transitions are managed by the \textit{ThoughtAnalyzerAgent}, which monitors progress and determines whether to continue, complete, or discard a branch. This lifecycle ensures that only relevant outputs contribute to the final analysis while preserving an auditable trace of reasoning paths.

\section{Systems Compared}
\label{appendix:systems-compared}

We compare the reported systems under the same reduced-corpus evidence restriction. All agentic systems use GPT-4.1 as the underlying LLM.

\paragraph{\textsc{Litmus (Re)Agent} (ours).}
Our full system answers each query using a dynamic DAG of specialised Thought Agents together with the three improvements introduced in Section~\ref{sec:improvements}: enriched expert knowledge, enhanced code execution with linguistic feature libraries, and prompt stabilisation. Queries are decomposed into hypothesis-specific branches that are investigated in parallel and then aggregated.

\paragraph{LITMUS++ (prior DAG system).}
The predecessor system \cite{anonymous2025predictive} uses the same DAG-based architecture with multiple Thought Agents, but without the three improvements introduced in this work: the knowledge base is smaller, the Coder agent lacks access to \texttt{lang2vec} and URIEL features, and prompts follow the original design. Comparing LITMUS++ to \textsc{Litmus (Re)Agent} isolates the effect of the three proposed changes.

\paragraph{ThoughtAgent (single multi-agent group chat, no DAG).}
A single ThoughtAgent that operates as a multi-agent group chat of specialised sub-agents---Research Planner, Web Search and Crawl, Expert Knowledge, Coder, and Reporter---but without DAG-level decomposition into parallel hypotheses. It processes the full query in one thread through coordinated sub-agent interaction. This baseline isolates the contribution of the multi-agent group chat architecture when hypothesis parallelism is absent.

\paragraph{Single Agent (ReAct loop, no group chat).}
A single ReAct agent with direct access to code execution, web search, and expert knowledge base tools in one agent loop, without the multi-agent group chat or DAG decomposition. This is the simplest agentic baseline and isolates the value of tool-augmented reasoning without any multi-agent coordination.

\paragraph{GPT-4.1 Direct Baseline.}
This baseline answers each query in a single GPT-4.1 call using the available evidence provided in-context, without iterative tool use or agentic decomposition. It measures the value added by agentic scaffolding beyond direct prompting with access to the same restricted evidence.

\paragraph{Magentic-One \cite{fourney2024magentic}.}
A general-purpose multi-agent framework that coordinates multiple agents through a lead orchestrator agent. Unlike \textsc{Litmus (Re)Agent}, it is not designed specifically for multilingual evaluation and does not include domain-specific tools, linguistic feature libraries, or a curated knowledge base. Including Magentic-One tests whether a general-purpose multi-agent approach can match domain-specific design on this task.

\subsection{LLM Hyperparameters}
\label{appendix:hyperparameters}

All systems use GPT-4.1 via the Azure OpenAI API (version \texttt{2025-01-01-preview}). The main agentic systems (\textsc{Litmus (Re)Agent}, LITMUS++, ThoughtAgent, and Single Agent) use a default configuration with temperature 1.0 and top-$p$ 1.0. The GPT-4.1 direct baseline uses temperature 0.7 with a maximum output length of 2{,}000 tokens.

The Expert Knowledge (Expert KB) agent operates with temperature 0 for deterministic retrieval and is augmented with a web search database. Retrieved documents are filtered using cosine similarity with a threshold of 0.90, and the top-$k$ results ($k=2$) are selected. A caching mechanism is used across experiments for efficiency. For structured output parsing, a separate LLM call with higher token limits and fallback handling is used. All evaluation is performed using GPT-4.1 as the LLM judge.

\section{Implementation Details}
\label{appendix:implementation}

\textbf{Tooling.} The backend orchestration uses \textit{Autogen} for multi-agent coordination, \textit{ChromaDB} as the vector store for the curated knowledge base, and the Firecrawl API for web search and scraping. ThoughtAgents invoke modular tools for search, scraping, PDF/JSON parsing, knowledge-base retrieval, and lightweight regression.

\textbf{Deployment.} The system runs locally or via a hosted interface, including a Dockerised setup for reproducibility. External search requires a user-provided API key, while other components operate over the benchmark corpus.

\textbf{Outputs.} Final predictions are returned together with supporting citations and structured reasoning traces. When applicable, the system also returns an uncertainty estimate produced during the aggregation stage.

\textbf{Extensibility.} The DAG orchestration is model- and task-agnostic: adding new tools or agents requires registration in the orchestration graph without changing the overall control flow. Related details on agent lifecycle and knowledge-base curation are provided in Appendices~\ref{appendix:agent-lifecycle} and~\ref{appendix:knowledge-base}.

\section{Knowledge Base Curation}
\label{appendix:knowledge-base}

A curated multilingual knowledge base supports \textsc{Litmus (Re)Agent} during hypothesis formation and feature-informed analysis. It integrates (i) \textbf{literature-derived resources} from peer-reviewed papers, benchmarks, and typological databases \cite{lauscher-etal-2020-zero, dolicki2021analysing, srinivasan2022litmus, ahuja2022beyond, kumar2023ditto}, and (ii) structured expert-oriented notes used to guide analysis strategies in sparse-evidence settings. The knowledge base is organised around language--task--model relations together with reusable analytical patterns, such as common transfer assumptions, feature types, and failure modes encountered in multilingual evaluation. In the current version, the knowledge base is used as structured guidance rather than as ground truth itself.

\section{Detailed Task--Scenario Results}
\label{appendix:task-scenario}

Table~\ref{tab:task_scenario_mae} reports PredSet MAE, QnASet accuracy, and LLM-judge quality scores for each task across all six systems.

\begin{table*}[htbp]
\centering
\setlength{\tabcolsep}{1.8pt}
\renewcommand{\arraystretch}{1.08}
\begin{tabular}{ll | rrrrr r | rrrrr r | cccc c}
\hline
 & & \multicolumn{6}{c|}{\textbf{PredSet MAE} ($\downarrow$)} & \multicolumn{6}{c|}{\textbf{QnASet Accuracy \%} ($\uparrow$)} & \multicolumn{5}{c}{\textbf{Quality (1--5)} ($\uparrow$)} \\
\textbf{Task} & \textbf{Sys} & \textbf{S1} & \textbf{S2} & \textbf{S3} & \textbf{S4} & \textbf{S5} & \textbf{Avg} & \textbf{S1} & \textbf{S2} & \textbf{S3} & \textbf{S4} & \textbf{S5} & \textbf{Avg} & \textbf{Pr} & \textbf{Ft} & \textbf{Co} & \textbf{Ci} & \textbf{Avg} \\
\hline
\multirow{6}{*}{\rotatebox{0}{Code Gen.}}
 & LRA & \textbf{7.9} & \textbf{14.4} & \textbf{8.5} & 20.2 & 20.4 & \textbf{14.2} & \textbf{88.0} & 32.0 & \textbf{73.9} & 38.1 & \textbf{40.0} & \textbf{55.3} & \textbf{4.7} & \textbf{4.9} & 4.9 & 3.0 & \textbf{4.4} \\
 & L++ & 10.5 & 29.9 & 22.7 & \textbf{18.4} & 24.6 & 21.1 & 32.0 & 16.0 & 34.8 & 19.0 & 25.0 & 25.4 & 3.1 & 3.3 & 4.6 & 2.1 & 3.3 \\
 & TA  & 16.8 & 23.6 & 12.0 & 24.7 & 18.6 & 18.9 & 60.0 & 20.0 & 43.5 & 42.9 & 30.0 & 39.5 & 4.6 & 4.7 & 4.8 & 3.0 & 4.3 \\
 & SA  & 12.0 & 23.5 & 12.0 & 21.8 & 21.1 & 17.8 & 76.0 & \textbf{48.0} & 65.2 & \textbf{47.6} & 15.0 & 51.8 & 4.4 & 3.9 & 4.9 & 4.2 & 4.4 \\
 & GPT & 28.4 & 43.7 & 39.5 & 47.7 & 39.8 & 39.8 & 24.0 & 20.0 & 43.5 & 23.8 & 25.0 & 27.2 & 4.4 & 3.3 & \textbf{5.0} & \textbf{4.6} & 4.3 \\
 & M1  & 17.8 & 30.9 & 22.4 & 38.7 & \textbf{26.0} & 27.0 & 84.0 & 24.0 & 69.6 & 9.5 & 10.0 & 41.2 & 3.8 & 3.0 & 4.9 & 4.2 & 4.0 \\
\hline
\multirow{6}{*}{\rotatebox{0}{Math Reas.}}
 & LRA & \textbf{14.0} & 17.6 & 26.2 & 26.6 & 14.8 & 20.0 & 12.0 & \textbf{24.0} & 25.0 & \textbf{12.5} & 4.0 & 15.5 & \textbf{4.6} & \textbf{4.9} & 4.9 & 3.0 & \textbf{4.3} \\
 & L++ & 28.0 & 23.7 & 31.7 & 26.7 & 32.5 & 28.8 & \textbf{16.0} & 8.0 & 12.5 & \textbf{12.5} & 4.0 & 10.6 & 3.2 & 3.5 & 4.7 & 2.2 & 3.4 \\
 & TA  & 25.5 & 25.0 & 27.2 & 41.4 & 19.6 & 27.5 & \textbf{32.0} & 8.0 & 16.7 & 8.3 & 0.0 & 13.0 & 4.5 & 4.6 & 4.7 & 2.8 & 4.2 \\
 & SA  & 30.0 & 22.8 & 29.0 & 29.8 & 36.4 & 29.6 & 4.0 & 12.0 & 16.7 & 4.2 & \textbf{12.0} & 9.8 & 4.3 & 3.9 & 4.9 & 4.2 & 4.3 \\
 & GPT & 18.0 & \textbf{6.7} & \textbf{23.3} & \textbf{20.8} & \textbf{12.3} & \textbf{18.2} & 12.0 & \textbf{24.0} & \textbf{29.2} & \textbf{16.7} & \textbf{12.0} & \textbf{18.7} & 4.1 & 2.8 & \textbf{5.0} & \textbf{4.0} & 4.0 \\
 & M1  & 23.0 & 15.4 & 28.5 & 37.9 & 22.1 & 26.8 & 12.0 & 4.0 & 12.5 & 4.2 & 4.0 & 7.3 & 3.7 & 2.8 & 4.8 & 3.9 & 3.8 \\
\hline
\multirow{6}{*}{\rotatebox{0}{QA/VQA}}
 & LRA & 13.4 & 11.7 & 7.2 & \textbf{5.7} & 8.7 & 9.5 & 0.0 & 0.0 & 28.0 & 4.0 & 8.0 & 8.0 & \textbf{4.5} & \textbf{4.8} & 4.8 & 2.9 & \textbf{4.2} \\
 & L++ & 14.9 & \textbf{6.7} & \textbf{3.6} & \textbf{4.7} & 14.7 & \textbf{8.7} & \textbf{20.0} & \textbf{16.0} & \textbf{24.0} & \textbf{32.0} & \textbf{20.0} & \textbf{22.4} & 4.2 & 4.6 & 4.9 & 2.7 & 4.1 \\
 & TA  & 23.3 & 14.6 & 12.3 & 10.9 & 10.9 & 14.7 & 20.0 & 4.0 & 12.0 & 4.0 & 0.0 & 8.0 & 4.1 & 4.2 & 4.3 & 2.7 & 3.8 \\
 & SA  & \textbf{12.0} & 18.2 & 11.6 & 9.0 & 11.6 & 12.4 & \textbf{24.0} & \textbf{20.0} & 8.0 & 12.0 & 20.0 & 16.8 & 4.2 & 3.8 & 4.7 & 4.0 & 4.2 \\
 & GPT & 13.0 & 8.8 & 8.4 & 6.9 & 11.1 & 9.6 & \textbf{20.0} & \textbf{16.0} & \textbf{24.0} & 4.0 & 16.0 & 16.0 & 4.1 & 2.7 & \textbf{5.0} & \textbf{4.1} & 4.0 \\
 & M1  & 17.8 & 13.7 & 12.0 & 12.3 & \textbf{13.7} & 14.1 & \textbf{20.0} & 4.0 & 4.0 & 0.0 & 0.0 & 5.6 & 3.7 & 3.1 & 4.7 & 3.6 & 3.8 \\
\hline
\multirow{6}{*}{\rotatebox{0}{Class./NLI}}
 & LRA & \textbf{8.7} & 8.6 & \textbf{3.2} & \textbf{4.8} & 13.5 & \textbf{7.9} & \textbf{36.0} & 8.0 & \textbf{16.0} & \textbf{12.0} & \textbf{8.0} & \textbf{16.0} & \textbf{4.6} & \textbf{4.7} & 4.9 & 3.0 & \textbf{4.3} \\
 & L++ & 14.1 & \textbf{8.8} & 5.5 & 12.3 & \textbf{8.3} & 9.6 & 8.0 & \textbf{16.0} & 0.0 & \textbf{12.0} & 0.0 & 7.2 & 3.1 & 3.4 & 4.6 & 2.4 & 3.4 \\
 & TA  & 15.0 & 7.9 & 5.5 & 7.9 & 17.7 & 11.6 & 8.0 & 4.0 & 12.0 & 4.0 & 4.0 & 6.4 & 4.2 & 4.2 & 4.3 & 2.8 & 3.9 \\
 & SA  & 14.7 & 14.9 & 5.8 & 10.3 & 10.4 & 11.9 & 8.0 & \textbf{20.0} & \textbf{20.0} & \textbf{16.0} & 4.0 & 13.6 & 4.3 & 3.6 & 4.7 & 4.2 & 4.2 \\
 & GPT & 12.2 & \textbf{8.8} & 3.4 & 10.9 & 13.5 & 9.9 & 8.0 & 8.0 & \textbf{20.0} & \textbf{12.0} & 0.0 & 9.6 & 4.3 & 2.5 & \textbf{5.0} & \textbf{4.5} & 4.1 \\
 & M1  & 14.8 & 14.9 & 5.7 & 12.5 & 11.0 & 12.0 & 8.0 & 0.0 & 4.0 & 0.0 & 0.0 & 2.4 & 3.9 & 3.2 & 4.8 & 3.9 & 4.0 \\
\hline
\multirow{6}{*}{\rotatebox{0}{Text Sum.}}
 & LRA & 2.5 & 6.1 & 2.7 & 8.4 & 14.2 & \textbf{6.5} & 8.0 & 16.0 & 36.0 & 16.0 & 31.6 & 21.0 & \textbf{4.5} & \textbf{4.6} & 4.9 & 2.9 & \textbf{4.2} \\
 & L++ & 4.7 & 18.1 & 3.1 & 11.3 & 23.3 & 12.6 & 4.0 & 8.0 & \textbf{44.0} & 0.0 & 10.5 & 13.4 & 3.3 & 3.4 & 4.6 & 2.5 & 3.4 \\
 & TA  & 4.7 & 14.8 & \textbf{2.0} & 12.4 & 15.8 & 9.8 & 0.0 & 12.0 & 20.0 & 8.0 & 15.8 & 10.9 & 4.4 & 4.4 & 4.6 & 2.9 & 4.1 \\
 & SA  & \textbf{2.2} & \textbf{2.3} & 3.8 & \textbf{6.7} & 20.1 & 7.1 & \textbf{12.0} & \textbf{28.0} & \textbf{44.0} & 8.0 & 15.8 & \textbf{21.8} & 4.0 & 3.5 & 4.5 & 3.8 & 4.0 \\
 & GPT & 2.9 & 14.0 & \textbf{1.9} & 15.6 & 17.0 & 9.6 & 4.0 & 8.0 & 40.0 & 4.0 & 21.1 & 15.1 & 4.2 & 2.8 & \textbf{5.0} & \textbf{4.6} & 4.1 \\
 & M1  & 3.7 & 12.2 & 2.8 & 15.9 & \textbf{14.2} & 9.2 & 4.0 & 4.0 & 0.0 & 0.0 & 0.0 & 1.7 & 3.8 & 3.0 & 4.8 & 4.0 & 3.9 \\
\hline
\multirow{6}{*}{\rotatebox{0}{Machine Tr.}}
 & LRA & 9.4 & 5.6 & 8.5 & \textbf{5.1} & \textbf{4.1} & \textbf{6.2} & \textbf{32.0} & 8.0 & \textbf{24.0} & 4.0 & \textbf{16.0} & 16.8 & \textbf{4.7} & \textbf{4.8} & 4.9 & 3.1 & \textbf{4.4} \\
 & L++ & \textbf{4.4} & 6.9 & \textbf{5.8} & 7.2 & 6.6 & 6.4 & 24.0 & 8.0 & 12.0 & 4.0 & 12.0 & 12.0 & 3.2 & 3.4 & 4.6 & 2.6 & 3.5 \\
 & TA  & 14.2 & 10.2 & 10.6 & 8.2 & 7.9 & 9.9 & \textbf{40.0} & 8.0 & 16.0 & \textbf{8.0} & \textbf{24.0} & \textbf{19.2} & 4.6 & 4.6 & 4.8 & 3.2 & 4.3 \\
 & SA  & 8.2 & 7.3 & 9.2 & 6.9 & 6.7 & 7.5 & 12.0 & 4.0 & 16.0 & 0.0 & 16.0 & 9.6 & 4.0 & 3.4 & 4.6 & 4.0 & 4.0 \\
 & GPT & 9.5 & 8.9 & 9.3 & 7.3 & 7.4 & 8.3 & 4.0 & 0.0 & \textbf{24.0} & \textbf{8.0} & 12.0 & 9.6 & 4.1 & 2.7 & \textbf{5.0} & \textbf{4.3} & 4.0 \\
 & M1  & 9.8 & 8.8 & 8.0 & 13.7 & 9.1 & 10.0 & 8.0 & 4.0 & 0.0 & 0.0 & 8.0 & 4.0 & 3.9 & 3.1 & 4.9 & 4.5 & 4.1 \\
\hline
\end{tabular}
\caption{Detailed results by task: PredSet MAE ($\downarrow$), QnASet accuracy \% ($\uparrow$), and LLM-judge quality (1--5, $\uparrow$). Quality: Pr = Predictive Plausibility, Ft = Feature Selection, Co = Coherence, Ci = Citation Emphasis. Abbreviations as in Table~\ref{tab:task_results_combined}.}
\label{tab:task_scenario_mae}
\end{table*}

\subsection{Per-Metric MAE Analysis}

Figure~\ref{fig:per_metric_mae} shows the mean absolute error of \textsc{Litmus (Re)Agent} broken down by evaluation metric type. Accuracy-based metrics (used primarily in code generation, classification/NLI, and mathematical reasoning) exhibit the highest MAE (12.7), reflecting the wider score ranges and greater prediction difficulty in these tasks. In contrast, text-overlap metrics like ROUGE (6.5), BLEU (5.0), and chrF (7.5), show substantially lower MAE. This pattern is consistent with the per-task results in Table~\ref{tab:task_results_combined}: tasks evaluated with accuracy metrics tend to produce more variable scores, while summarisation and translation metrics operate on narrower, more predictable ranges. These results suggest that metric-aware normalisation or task-specific calibration may further improve prediction quality.

\begin{figure}[htbp]
    \centering
    \includegraphics[width=0.48\textwidth]{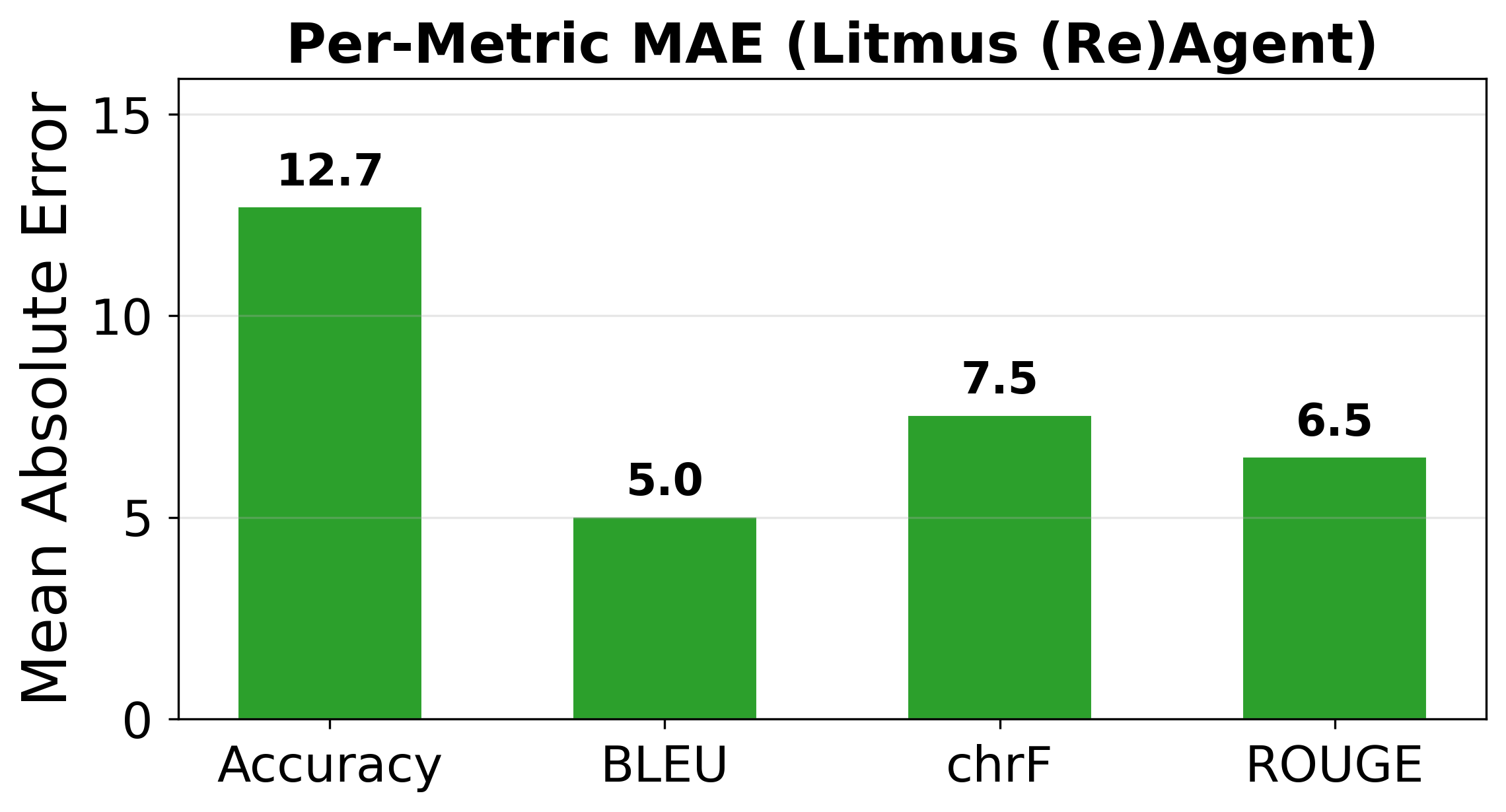}
    \caption{Per-metric MAE for \textsc{Litmus (Re)Agent}. Accuracy-based metrics show higher prediction error than text-overlap metrics such as ROUGE, BLEU, and chrF.}
    \label{fig:per_metric_mae}
\end{figure}

\subsection{Per-Scenario MAE Trends}

Figure~\ref{fig:per_scenario_line} presents the per-scenario MAE as a line chart, showing how each system's prediction error evolves across the five evidence configurations. Several patterns are visible. First, \textsc{Litmus (Re)Agent} maintains the lowest MAE across all scenarios, with a relatively flat trajectory (range 9.0--11.9). Second, all systems show increased MAE in S4 (distant language transfer), but this effect is most pronounced for Magentic-One (23.5), which lacks the linguistic feature tools needed for cross-lingual reasoning. Third, S3 (similar language transfer) often yields lower MAE than S1 (direct evidence), confirming the non-monotonic difficulty pattern discussed in Section~\ref{sec:results}. Fourth, the gap between \textsc{Litmus (Re)Agent} and the other systems widens in S4, indicating that the system's typological feature extraction and regression capabilities provide the largest benefit under distant transfer.

\begin{figure}[htbp]
    \centering
    \includegraphics[width=0.48\textwidth]{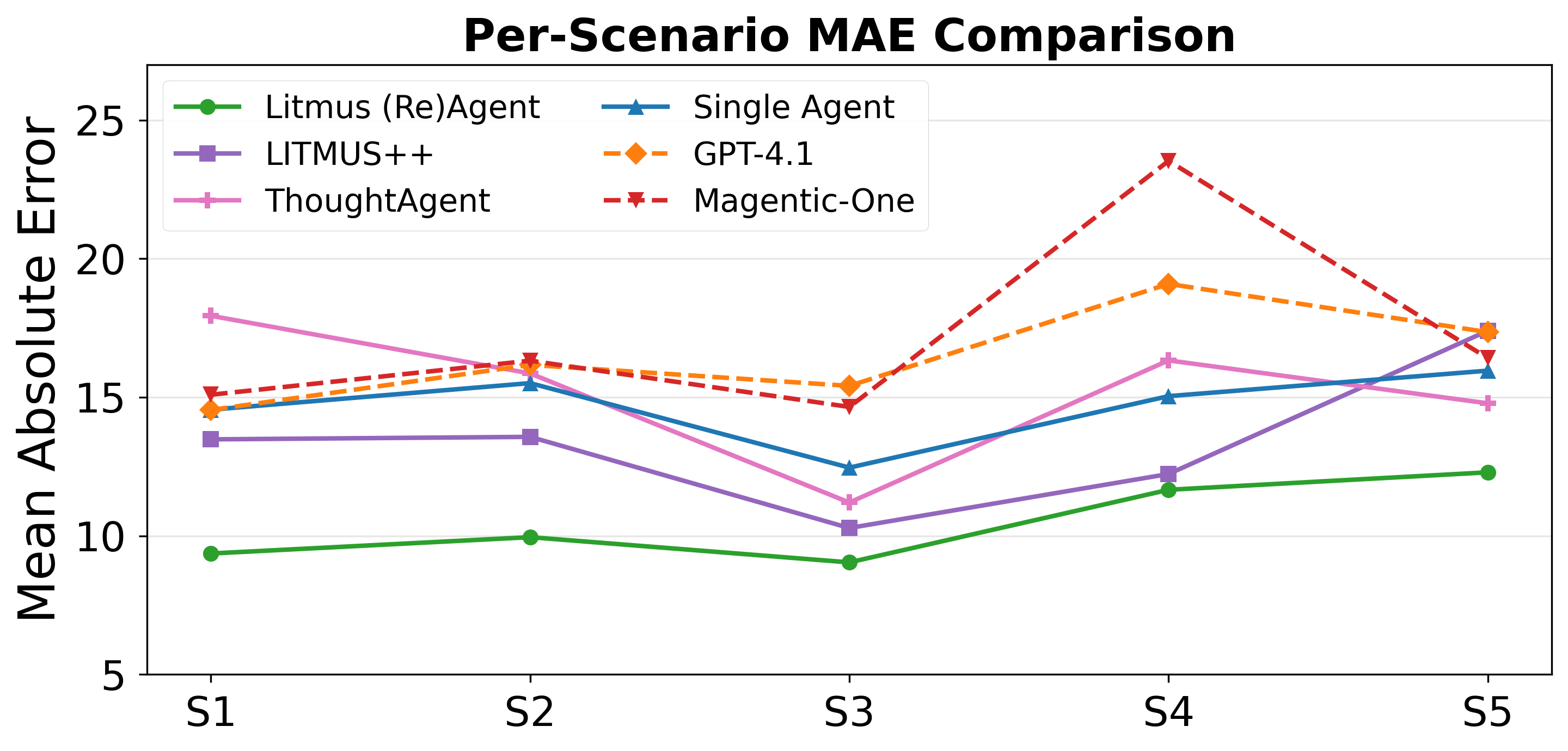}
    \caption{Per-scenario MAE trends across all five systems. \textsc{Litmus (Re)Agent} maintains the lowest and most stable MAE, while Magentic-One degrades sharply in S4 (distant language transfer).}
    \label{fig:per_scenario_line}
\end{figure}

\subsection{Detailed Error Analysis by Task and Scenario}

Figures~\ref{fig:mae_by_scenario_task} and~\ref{fig:mae_distribution_boxplot} provide a finer-grained analysis of \textsc{Litmus (Re)Agent}'s prediction errors across task--scenario combinations.

Figure~\ref{fig:mae_by_scenario_task} shows the mean MAE broken down by scenario (x-axis) with each task as a separate bar. Code generation and mathematical reasoning exhibit notably higher MAE than other tasks across most scenarios, consistent with the per-task results in Table~\ref{tab:task_results_combined}. Within code generation, S2 and S5 show elevated error, suggesting that cross-model and combined cross-lingual/cross-model transfer are particularly challenging for this task. In contrast, text summarisation and machine translation maintain consistently low MAE across all scenarios, indicating that the system's predictions are more stable for tasks evaluated with text-overlap metrics.

\begin{figure*}[htbp]
    \centering
    \includegraphics[width=\textwidth]{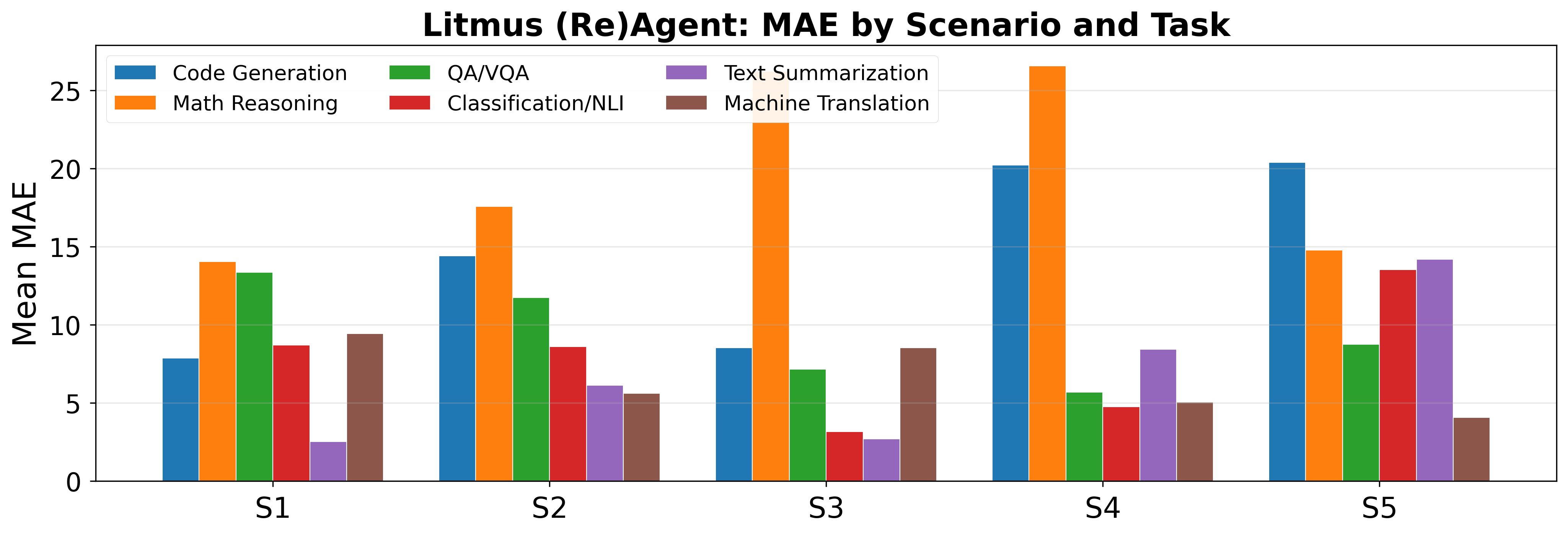}
    \caption{Litmus (Re)Agent: mean MAE by scenario and task. Each group of bars corresponds to one scenario (S1--S5), with tasks shown as different colours. Code generation and mathematical reasoning show higher and more variable error than other tasks.}
    \label{fig:mae_by_scenario_task}
\end{figure*}

Figure~\ref{fig:mae_distribution_boxplot} shows the full distribution of per-question absolute errors using box plots on a log scale. The distributions reveal that median errors are often low even when mean MAE is high, indicating that a small number of high-error predictions drive up the mean. This is most visible in code generation and mathematical reasoning under S4 and S5, where the interquartile range is wide and outliers are frequent. QA/VQA, classification/NLI, and machine translation show tighter distributions with fewer extreme errors. These patterns suggest that targeted improvements in high-error cases---particularly for code generation in distant-transfer scenarios---could yield substantial gains in overall MAE.

\begin{figure*}[htbp]
    \centering
    \includegraphics[width=\textwidth]{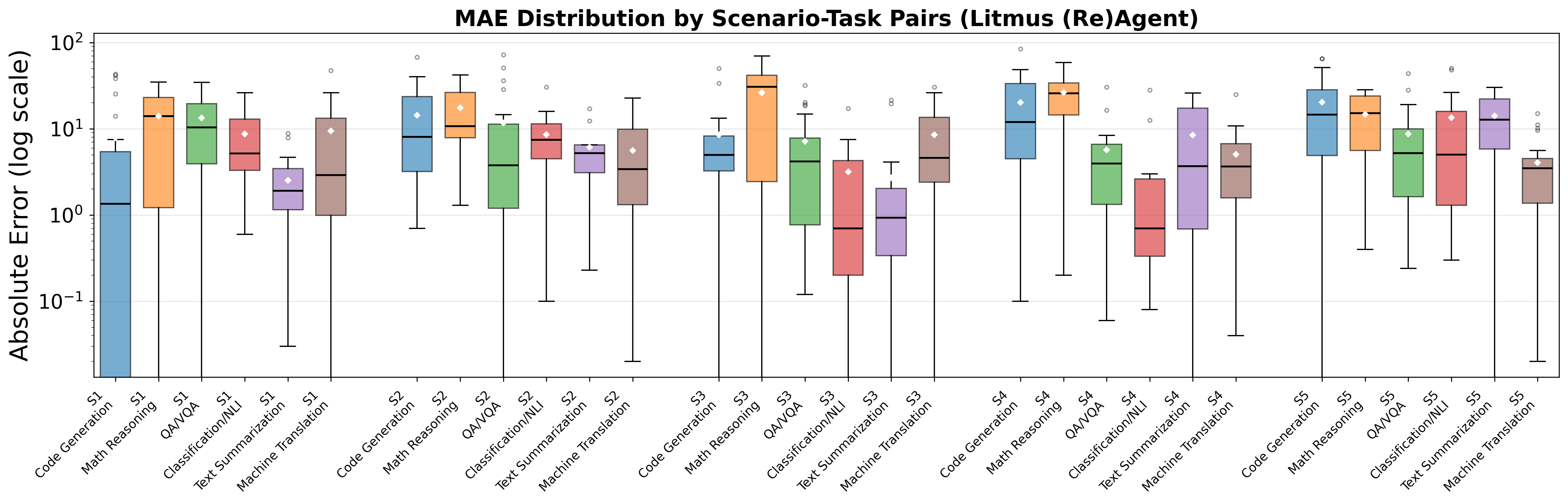}
    \caption{Distribution of per-question absolute errors for \textsc{Litmus (Re)Agent}, grouped by scenario--task pairs (log scale). Diamonds indicate means; horizontal lines indicate medians. Wider boxes and more outliers appear in code generation and math reasoning, especially in transfer-heavy scenarios.}
    \label{fig:mae_distribution_boxplot}
\end{figure*}




\section{Human Evaluation Details}
\label{appendix:human-eval-details}

This appendix provides additional analysis from the human evaluation study described in Section~\ref{sec:human_eval}.

\subsection{Participant Demographics}

The 8 participants comprise two groups. The \textbf{novice} group (4 participants) consists of undergraduate students in computer science or AI/ML programmes with less than one year of NLP evaluation experience and no prior exposure to multilingual benchmarking. The \textbf{expert} group (4 participants) includes a faculty member with a PhD in AI/NLP and 5+ years of experience, two researchers with 1--3 years of experience in code generation and speech/AI agents respectively, and a PhD student in NLP with 3--5 years of experience. Participants were recruited through direct invitation. Participation was voluntary and no monetary compensation was provided. All participants completed the baseline phase (5~questions each); most completed the \textsc{Litmus (Re)Agent} phase, though two experts answered fewer than 5 questions due to time constraints.

\subsection{Study Interface and Evaluation Form}

The human evaluation was conducted using \textsc{Litmus (Re)Agent}, an interactive tool built on top of the LITMUS++ benchmark. Figures~\ref{fig:ui_sample1} and~\ref{fig:ui_sample2} show representative screenshots of the tool interface. The system presents each prediction question along with a constrained set of accessible papers, and provides structured outputs including retrieved evidence, linguistic feature analysis, and regression-based predictions. Participants interact with the system in a conversational format and can request additional analysis or clarification.

\begin{figure}[htbp]
    \centering
    \includegraphics[width=0.48\textwidth]{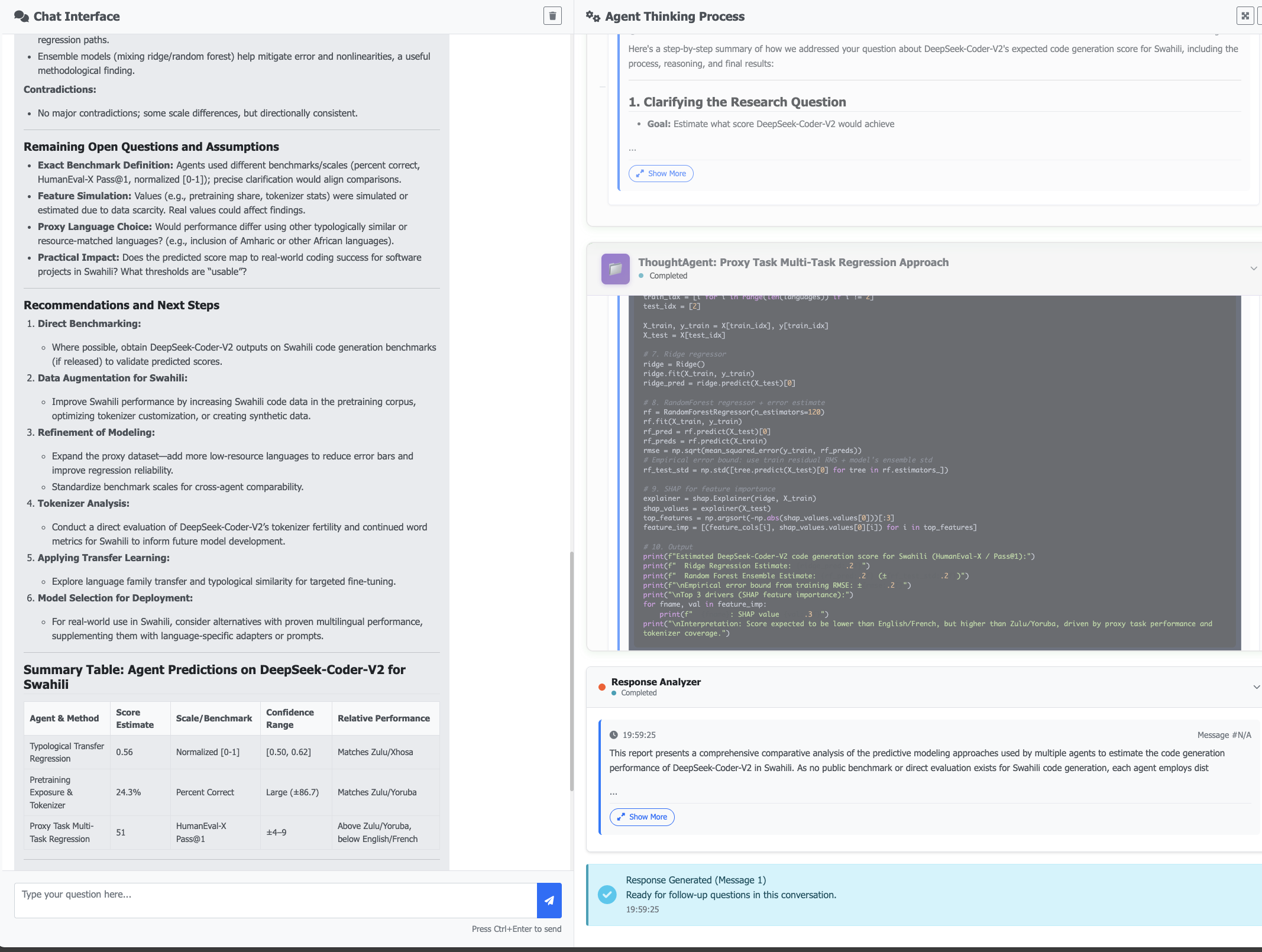}
    \caption{Sample screenshot of the \textsc{Litmus (Re)Agent} interface showing the conversational interaction, structured evidence retrieval, and prediction output presented to participants during the \textsc{Litmus (Re)Agent} phase.}
    \label{fig:ui_sample1}
\end{figure}

\begin{figure}[htbp]
    \centering
    \includegraphics[width=0.48\textwidth]{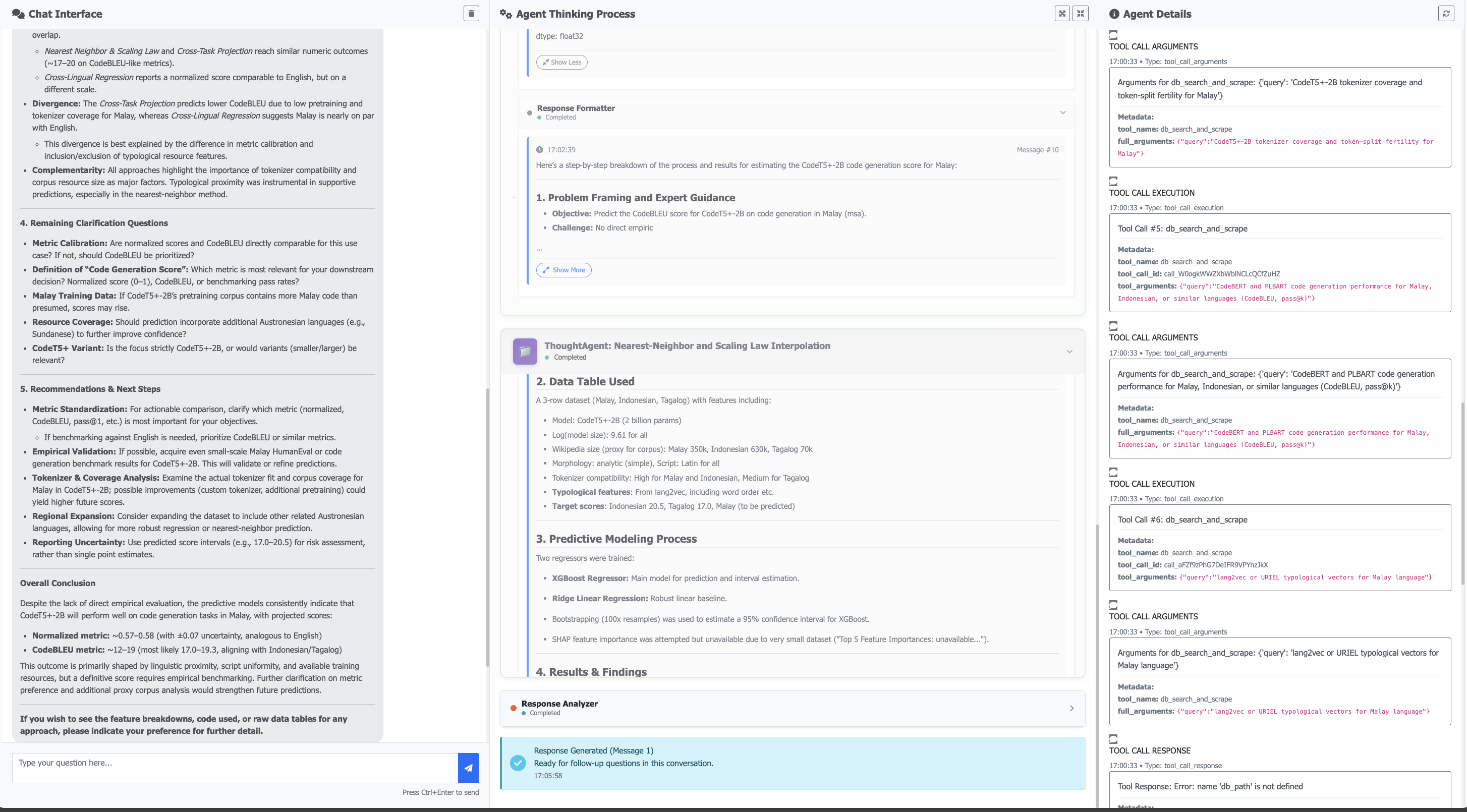}
    \caption{Sample screenshot of the \textsc{Litmus (Re)Agent} interface showing the final prediction summary with confidence intervals, reasoning trace, and supporting evidence.}
    \label{fig:ui_sample2}
\end{figure}

After each question, participants completed a structured evaluation form (Figure~\ref{fig:eval_form}). The form captures the predicted value, confidence interval, reasoning strategy, challenges encountered, and Likert-scale ratings on five quality dimensions (confidence, interpretability, explainability, justification, and actionability). Tool usefulness and effort saved ratings are also collected. This form was identical across both the baseline and \textsc{Litmus (Re)Agent} phases to ensure comparability. The complete list of evaluation questions, detailed metric rubrics, and per-section form structure are provided in the subsections below.

\begin{figure}[htbp]
    \centering
    \includegraphics[width=0.48\textwidth]{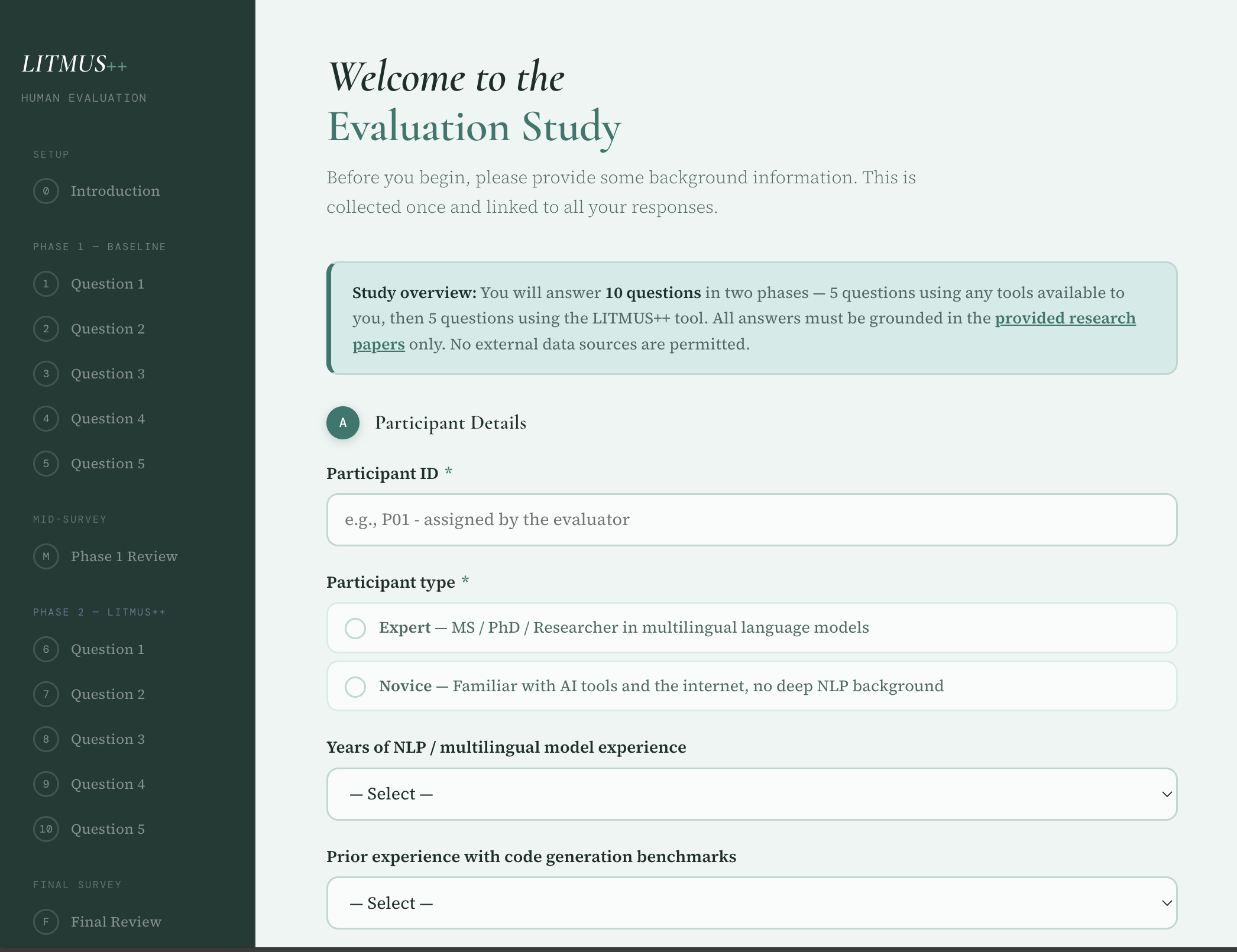}
    \caption{Screenshot of the human evaluation form used after each prediction question. Participants rate quality dimensions on Likert scales and report their reasoning strategies and challenges.}
    \label{fig:eval_form}
\end{figure}

\subsection{Evaluation Questions}

Table~\ref{tab:human_eval_questions} lists the 10 prediction questions used in the human evaluation study. All questions target code generation for low-resource or mid-resource languages, requiring participants to predict a numeric performance score (pass@1).

\begin{table}[htbp]
\centering
\footnotesize
\setlength{\tabcolsep}{2pt}
\begin{tabular}{c l l l}
\hline
\textbf{\#} & \textbf{Phase} & \textbf{Model} & \textbf{Language} \\
\hline
1 & Baseline & GPT-4 & Hindi \\
2 & Baseline & GPT-4o & Odia \\
3 & Baseline & GPT-3.5 & Romanian \\
4 & Baseline & DeepSeek-Coder-V2 & Shona \\
5 & Baseline & Claude-3.5-Opus & Bosnian \\
\hline
6 & LITMUS++ & CodeT5+-2B & Malay \\
7 & LITMUS++ & Claude-3.5-Opus & Macedonian \\
8 & LITMUS++ & GPT-4o & Bengali \\
9 & LITMUS++ & GPT-3.5 & Dari \\
10 & LITMUS++ & DeepSeek-Coder-V2 & Swahili \\
\hline
\end{tabular}
\caption{The 10 code generation prediction questions used in the human evaluation.}
\label{tab:human_eval_questions}
\end{table}

\subsection{Per-Question Form Structure}

For each question, participants completed a structured form (Figure~\ref{fig:eval_form}) with six sections:

\paragraph{Section A: Prediction.} A numeric predicted value (required) and an optional 80\% confidence interval (low--high range).

\paragraph{Section B: Reasoning Process.} Strategy used (multi-select from: direct lookup, interpolation, cross-paper synthesis, analogical reasoning, tool output, domain knowledge, educated guess, other), step-by-step reasoning (free text), papers referenced, and number of papers searched.

\paragraph{Section C: Challenges.} Multi-select from: missing data, inconsistent papers, too many papers, complex metric, tool limitation, unfamiliar language/model, no significant challenge.

\paragraph{Section D: Confidence and Interpretability.} Four Likert-scale ratings with detailed rubrics:

\begin{itemize}
\item \textbf{Confidence} (0--5): 0 = no basis, random estimate; 1 = vague intuition; 2 = partially relevant info, significant uncertainty; 3 = relevant patterns, not directly reported; 4 = closely matching data, minor extrapolation; 5 = exact or near-exact data point found.
\item \textbf{Interpretability} (1--5): 1 = gut feeling, cannot articulate steps; 3 = general approach clear, some steps fuzzy; 5 = every step explicit and traceable.
\item \textbf{Explainability} (1--5): 1 = cannot give coherent explanation; 3 = broad strokes OK, cannot justify specifics; 5 = fully transparent with paper references.
\item \textbf{Justification} (1--5): 1 = raw numbers, zero explanation; 3 = some factors explained, gaps remain; 5 = thorough, well-evidenced explanation.
\end{itemize}

\paragraph{Section E: Tool Experience.} Tool usefulness (1--5: 1 = no relevant info; 3 = useful direction, verification needed; 5 = directly provided the answer), effort saved (1--5: 1 = no savings or wasted time; 5 = tool essentially answered it), better/worse prediction vs.\ unaided (5-point scale from ``much worse'' to ``much better''), and optional free-text feedback.

\paragraph{Section F: Decision Quality.} Actionability (1--5: 1 = too uncertain for any decision; 3 = one input among many; 5 = sufficient for deployment decisions) and calibration surprise (how surprised if prediction wrong by $\pm$5 points, from ``not at all'' to ``extremely'').

\subsection{Mid-Survey and Final Survey}

Between phases, participants completed a mid-survey assessing Phase~1 cognitive load (mental demand, physical effort, frustration, success feeling) and tool experience (overall utility, ease of use). After Phase~2, a final survey collected: overall cognitive load for both phases, phase comparison (which produced better predictions, ease comparison, confidence comparison), LITMUS++ evaluation (utility, ease, transparency, trust, future use, comparison rank vs.\ other tools), highlights and failures (easiest/hardest questions, specific tool failures), and open feedback (most valuable aspect, most limiting, feature request).

\subsection{Quality Metric Profiles}

Table~\ref{tab:human_eval} provides the full numeric breakdown of human evaluation results, including novice--expert subgroup analysis.

\begin{table}[htbp]
\centering
\small
\resizebox{\columnwidth}{!}{
\begin{tabular}{l cc c cc c cc c}
\hline
 & \multicolumn{3}{c}{\textbf{Overall}} & \multicolumn{3}{c}{\textbf{Novice}} & \multicolumn{3}{c}{\textbf{Expert}} \\
\cmidrule(lr){2-4} \cmidrule(lr){5-7} \cmidrule(lr){8-10}
\textbf{Metric} & \textbf{B} & \textbf{L} & \textbf{$\Delta$} & \textbf{B} & \textbf{L} & \textbf{$\Delta$} & \textbf{B} & \textbf{L} & \textbf{$\Delta$} \\
\hline
Confidence       & 2.6 & 3.4 & \textbf{+.8} & 2.4 & 3.3 & \textbf{+1.0} & 2.8 & 3.5 & +.7 \\
Interpretability & 3.1 & 3.7 & \textbf{+.7} & 3.1 & 4.0 & \textbf{+.8} & 3.0 & 3.5 & +.5 \\
Explainability   & 3.2 & 3.9 & \textbf{+.7} & 3.2 & 4.0 & \textbf{+.8} & 3.4 & 3.9 & +.5 \\
Justification    & 2.8 & 3.8 & \textbf{+1.0} & 3.0 & 3.8 & +.9 & 2.8 & 3.8 & \textbf{+1.1} \\
Actionability    & 2.6 & 3.5 & \textbf{+.9} & 2.8 & 3.7 & \textbf{+1.0} & 2.5 & 3.3 & +.8 \\
\hline
Tool Useful.     & 3.5 & 3.8 & +.3 & 3.9 & 4.4 & +.5 & 3.1 & 3.1 & .0 \\
Effort Saved     & 3.7 & 3.9 & +.2 & 4.2 & 4.4 & +.2 & 3.3 & 3.1 & $-.1$ \\
Time (min)       & 8.6 & 12.0 & +3.5 & 5.9 & 13.7 & +7.7 & 10.7 & 10.3 & $-.3$ \\
\hline
\end{tabular}
}
\caption{Human evaluation: baseline (B) vs.\ \textsc{Litmus (Re)Agent} (L). Higher is better except for time.}
\label{tab:human_eval}
\end{table}

Figure~\ref{fig:human_eval_radar} shows the radar chart of all seven quality metrics broken down by participant type and phase. Novices show larger gains across most dimensions when moving from baseline to \textsc{Litmus (Re)Agent}, particularly in interpretability and actionability. Experts start from a higher baseline on confidence and explainability but show a pronounced improvement in justification quality. Tool usefulness and effort saved ratings remain relatively flat for experts, suggesting they derive value from the system's reasoning traces rather than from reduced manual effort.


\subsection{Self-Assessed Prediction Quality}

Figure~\ref{fig:human_eval_chart} shows quality dimensions and self-assessed prediction quality across the two phases. In the baseline phase, 60.0\% of responses were rated ``slightly better'' or ``much better'' than what participants felt they could achieve unaided; in the \textsc{Litmus (Re)Agent} phase, this proportion rose to 65.7\%. The distribution also shifted: ``much better'' responses increased from 15.0\% to 22.9\%, while ``same'' responses dropped from 22.5\% to 11.4\%.

\begin{figure}[htbp]
    \centering
    \includegraphics[width=0.48\textwidth]{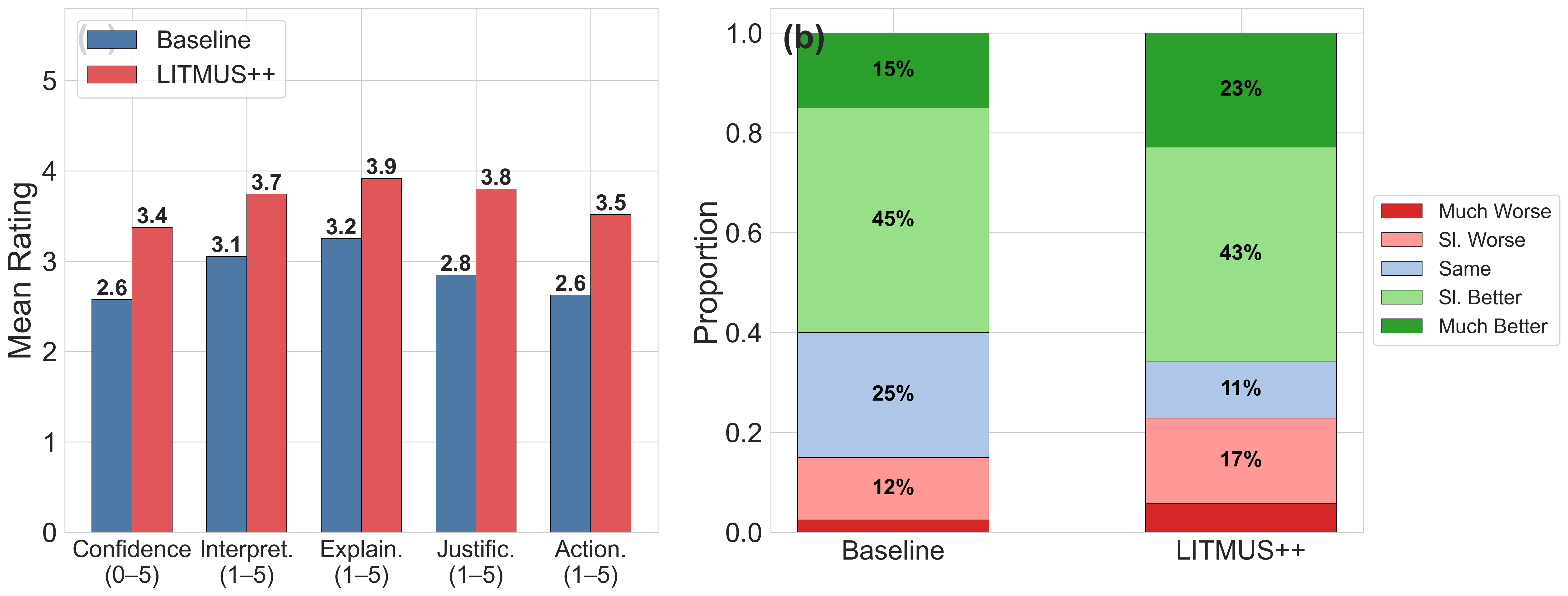}
    \caption{Human evaluation results: (a)~mean quality dimension ratings (baseline vs.\ \textsc{Litmus (Re)Agent}) across all participants, (b)~self-assessed prediction quality distribution showing the proportion of responses rated from ``much worse'' to ``much better'' relative to unaided prediction.}
    \label{fig:human_eval_chart}
\end{figure}

\subsection{Strategy Shifts}

In the baseline phase, participants relied on diverse strategies: interpolation (18), cross-paper search (14), tool output (14), analogical reasoning (11), and guessing (4). In the LITMUS++ phase, \textit{tool-output} became the dominant strategy (23 uses), while guessing nearly disappeared (4$\to$1). This shift reflects participants' reliance on the system's structured outputs rather than manual heuristics.

\subsection{Challenge Distribution}

The most common challenge across both phases was \textit{missing-data} (31 baseline, 27 LITMUS++). LITMUS++ reduced the prevalence of \textit{too-many} papers to process (10$\to$5) and \textit{tool-limitation} challenges (5$\to$1), suggesting that the system's automated retrieval and synthesis pipeline alleviates manual search burden.

\subsection{Calibration Confidence}

Participants were asked how surprised they would be if the true answer differed from their prediction by $\pm$5 percentage points. In the baseline phase, 25.0\% reported ``not at all'' surprised, indicating low confidence in their predictions. In the LITMUS++ phase, this dropped to 8.6\%, while ``very'' surprised rose from 7.5\% to 25.7\%, indicating that the system improved participants' calibration confidence.

\subsection{Workload and Final Survey}

A mid-survey (after baseline) and final survey (after LITMUS++) collected NASA-TLX-style workload ratings. Mental demand (3.4 vs.\ 3.4), frustration (3.5 vs.\ 3.6), and success feeling (2.8 vs.\ 2.7) remained similar across phases, while physical effort increased slightly (3.2$\to$3.6). In the final survey, participants rated the system on transparency (4.3/5), overall utility (3.9/5), ease of use (3.9/5), trust (3.7/5), and future use (3.7/5). Five participants indicated that LITMUS++ improved accuracy (3 ``clearly,'' 2 ``slightly''), while two rated it as equivalent.

\subsection{Qualitative Feedback}

The most valued aspects were deeper analysis with predictive models, use of linguistic features for regression, and clearer justification of predictions. The most common criticism was response latency (mentioned by 4 participants). One expert noted that LITMUS++ sometimes produced outputs that differed from other tools, requiring additional paper verification. Feature requests included faster response times, source hyperlinks, and broader general knowledge.

\section{Coder Agent Analysis}
\label{appendix:coder-analysis}

Figure~\ref{fig:algo_heatmap} shows the regression algorithms selected by the Coder agent across tasks. The dominant algorithms are Random Forest, Ridge Regression, and XGBoost, which together account for the majority of model-fitting attempts. The algorithm mix varies by task: code generation and math reasoning favour ensemble methods, while text summarisation and machine translation show higher usage of linear models. This variation reflects the Coder agent's ability to adapt its modelling strategy to the characteristics of each task.

\begin{figure}[htbp]
    \centering
    \includegraphics[width=0.48\textwidth]{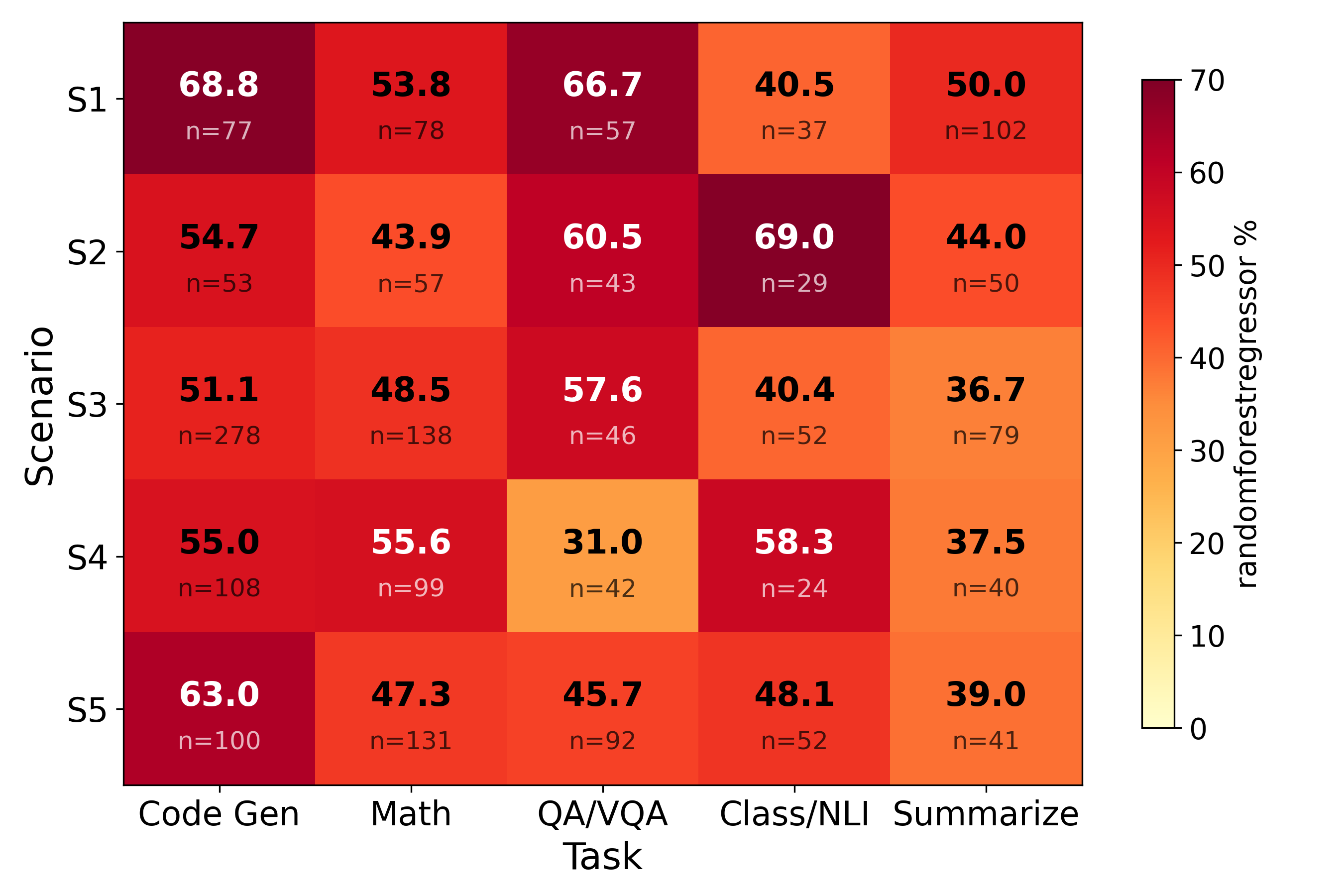}
    \caption{Regression algorithms selected by the Coder agent, broken down by task. Darker cells indicate higher usage frequency.}
    \label{fig:algo_heatmap}
\end{figure}

Figure~\ref{fig:lang2vec_scenario} shows how \texttt{lang2vec} feature usage varies across scenarios. Usage is highest in Scenarios~3 and~4 (similar and distant language transfer), where typological features are most informative for cross-lingual regression. In Scenario~1, where direct evidence is available, the agent relies less on typological features and more on retrieved numeric results.

\begin{figure}[htbp]
    \centering
    \includegraphics[width=0.48\textwidth]{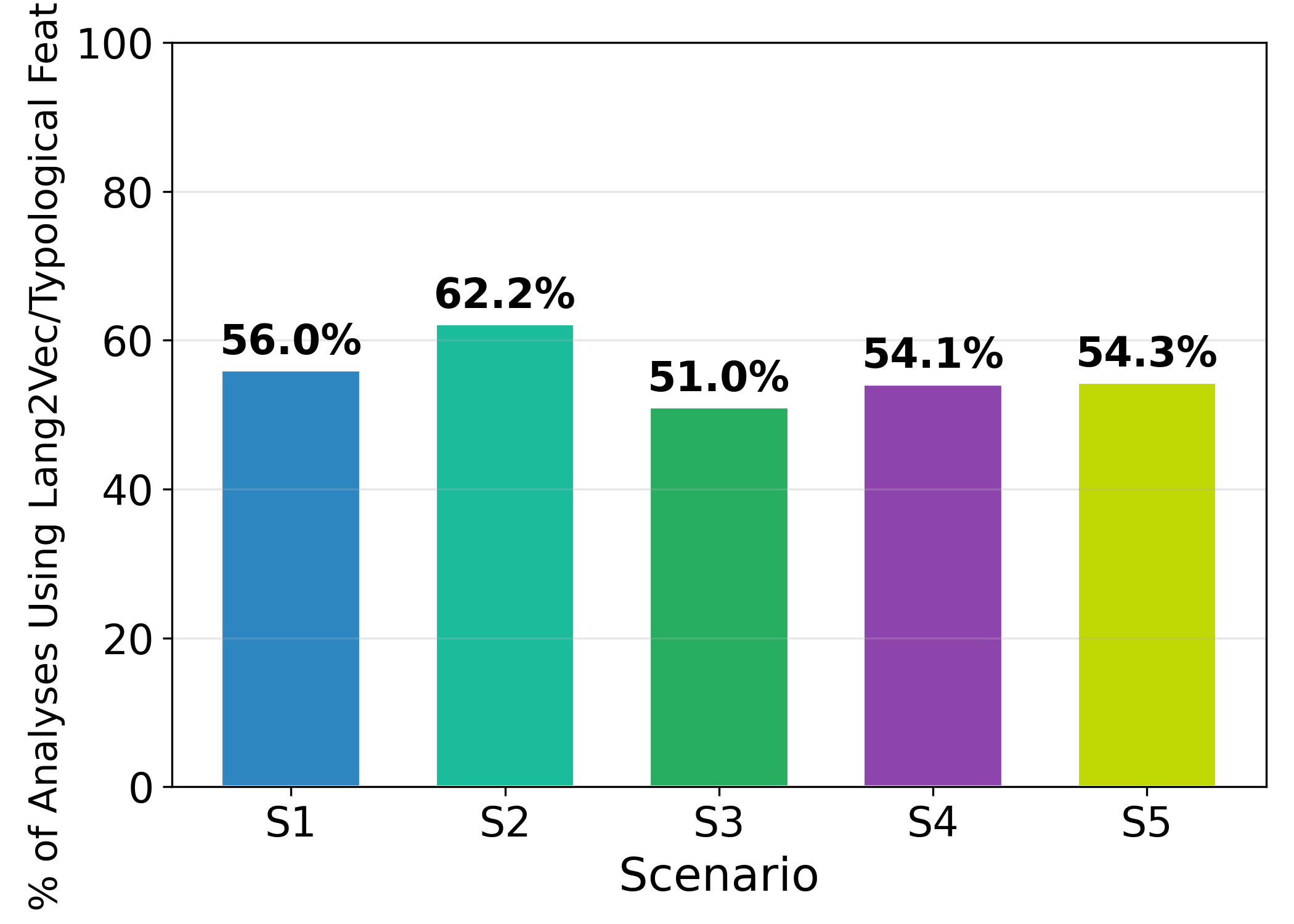}
    \caption{Usage of \texttt{lang2vec} typological features by scenario. Cross-lingual scenarios (S3, S4) show the highest feature utilisation.}
    \label{fig:lang2vec_scenario}
\end{figure}

Figure~\ref{fig:lang2vec_breakdown} provides a finer breakdown of \texttt{lang2vec} feature categories used by the Coder agent across tasks. Syntactic and geographic features are selected most frequently, with moderate use of phonological and genealogical signals. This pattern is consistent with feature-selection behaviour observed in cross-lingual transfer settings, where structural and areal similarity often provide stronger predictive cues than lexical overlap alone.

\begin{figure}[htbp]
    \centering
    \includegraphics[width=0.48\textwidth]{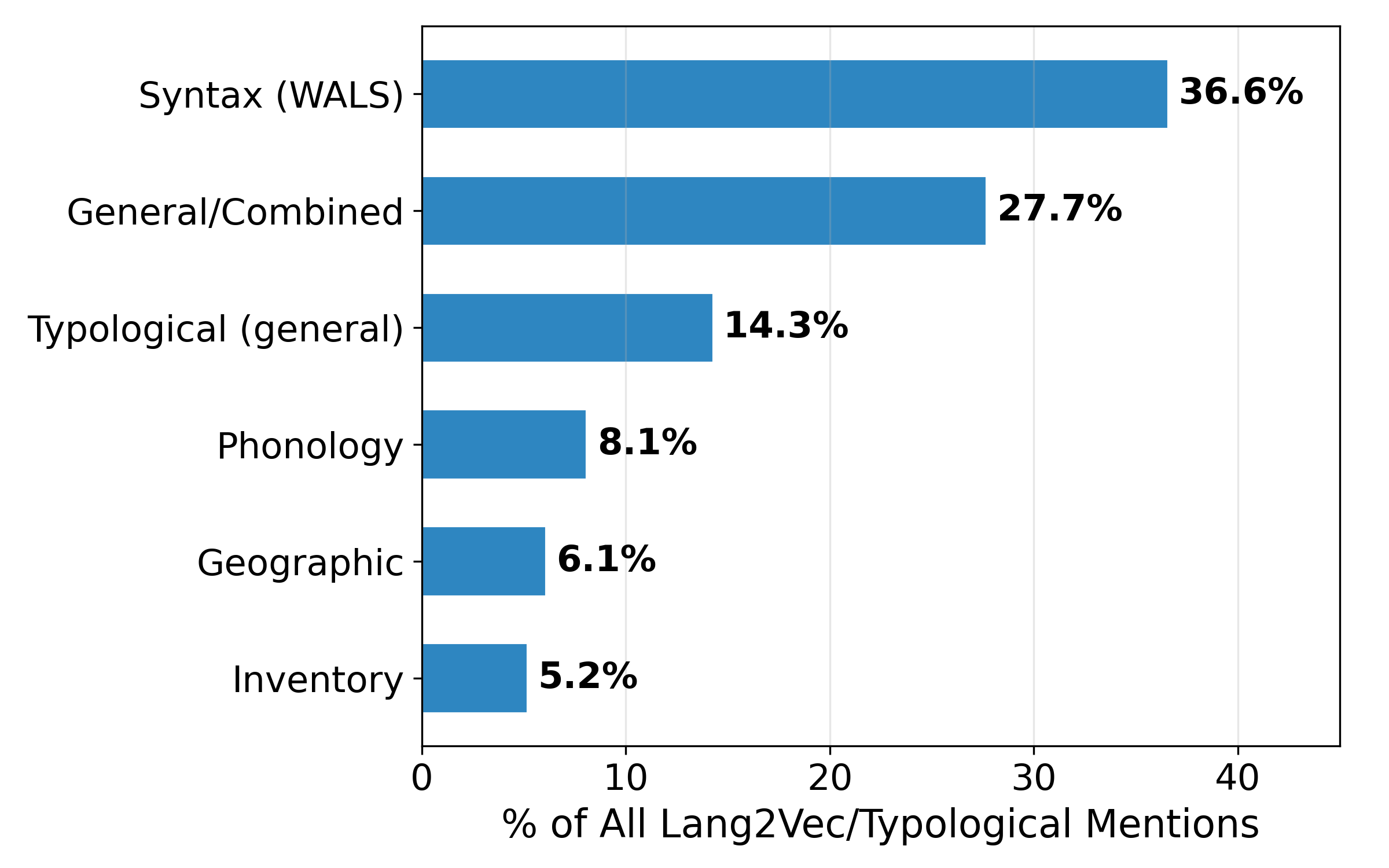}
    \caption{Breakdown of \texttt{lang2vec} feature categories used by the Coder agent. Syntactic and geographic features dominate across all tasks.}
    \label{fig:lang2vec_breakdown}
\end{figure}

\section{Benchmark Creation}
\label{appendix:benchmark-creation}

\subsection{Scenario-wise Coverage}

Figure~\ref{fig:scenario_lang_model_stats} reports scenario-wise benchmark coverage used for question construction. The left panel shows the number of unique languages represented in each task--scenario block, while the right panel shows the number of unique model families. As expected, direct-evidence scenarios provide broader coverage, while transfer-based scenarios operate under narrower and more selective support.

\begin{figure*}[htbp]
    \centering
    \begin{subfigure}[htbp]{0.48\textwidth}
        \centering
        \includegraphics[width=\textwidth]{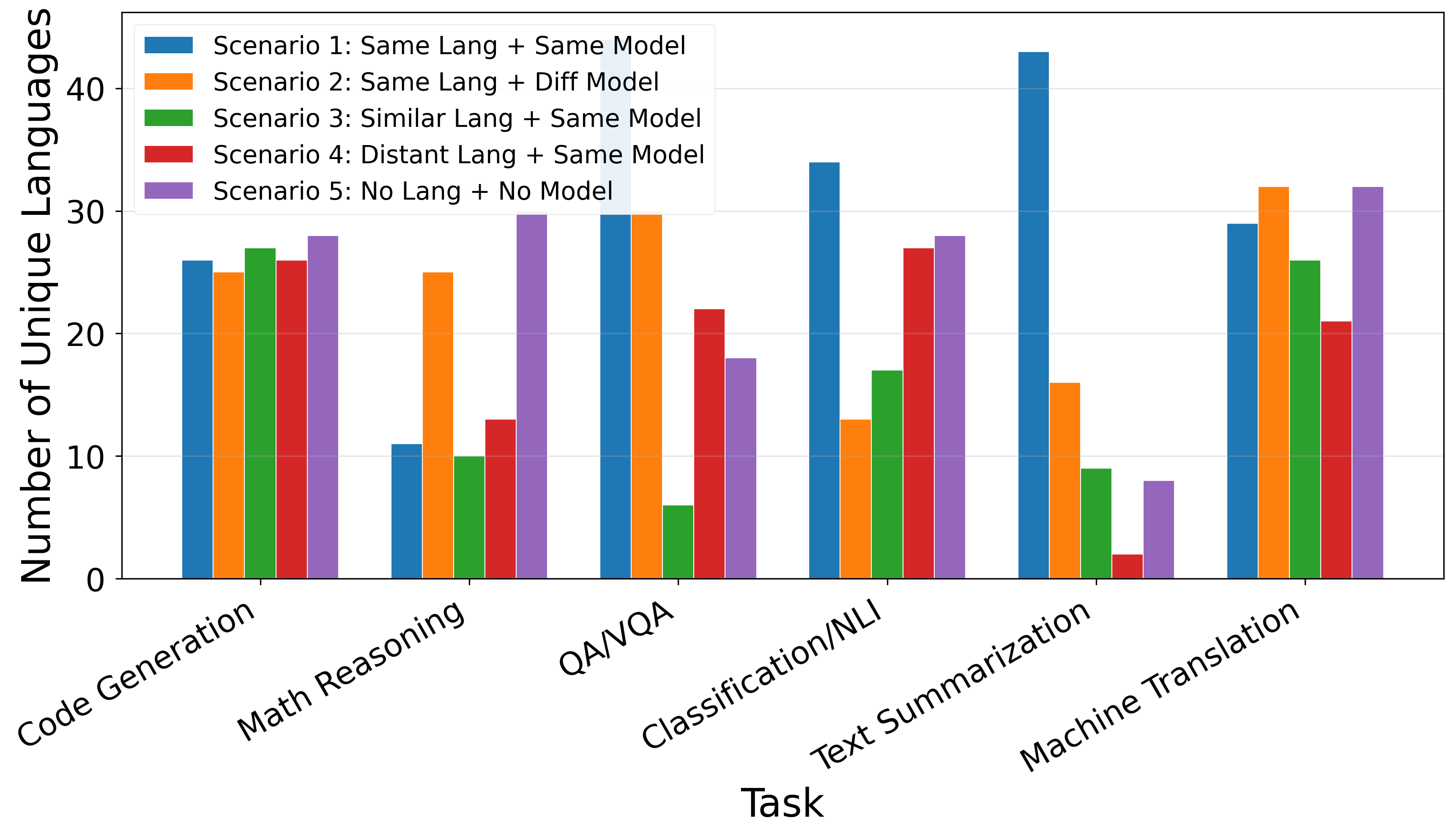}
        \caption{Number of unique languages used for question construction per task and scenario.}
        \label{fig:scenario_lang_stats}
    \end{subfigure}
    \hfill
    \begin{subfigure}[htbp]{0.48\textwidth}
        \centering
        \includegraphics[width=\textwidth]{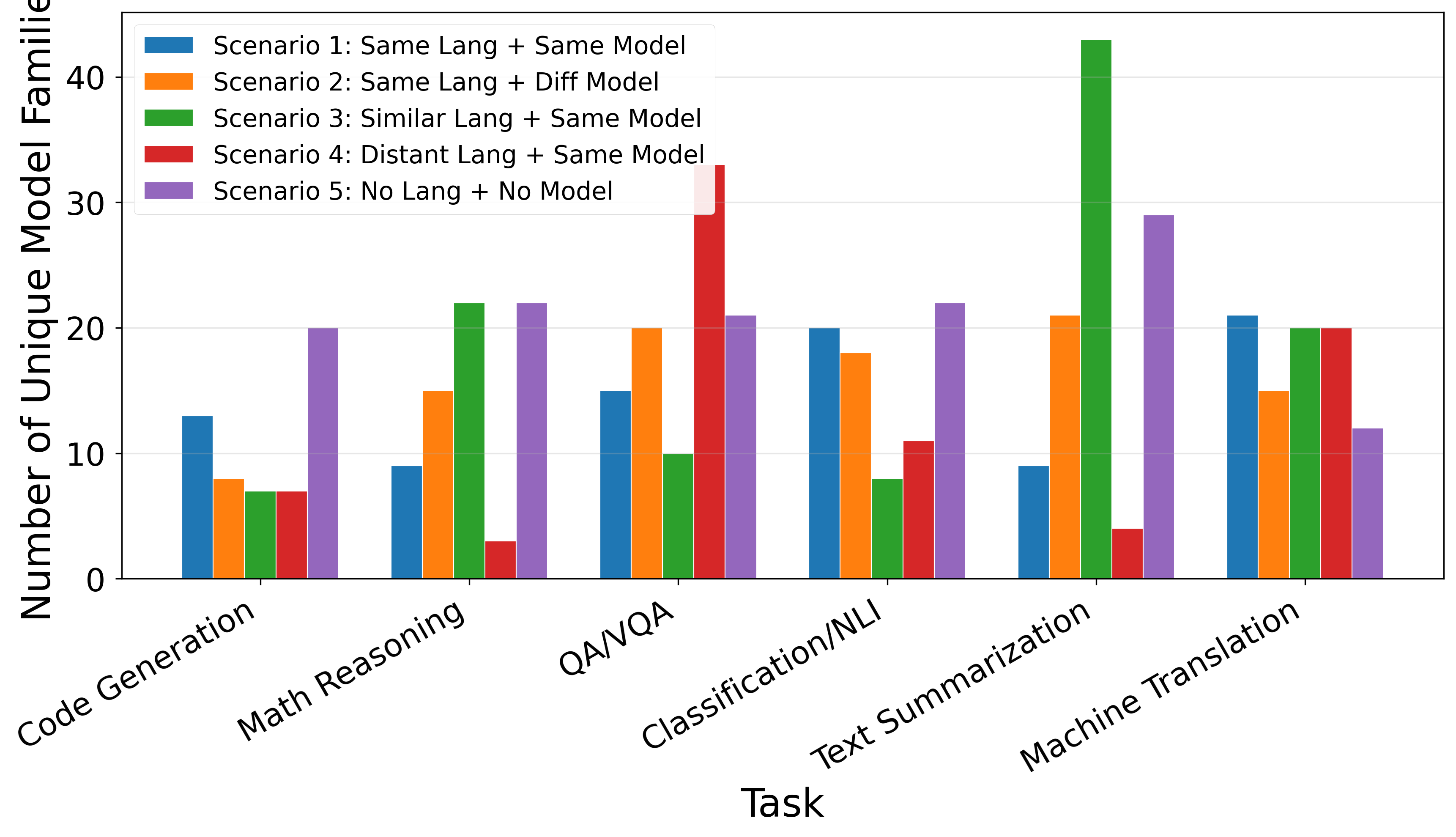}
        \caption{Number of unique model families used for question construction per task and scenario.}
        \label{fig:scenario_model_stats}
    \end{subfigure}
    \caption{Unique language and model-family coverage by task and scenario.}
    \label{fig:scenario_lang_model_stats}
\end{figure*}

\subsection{Language Distribution Across Questions}

Figure~\ref{fig:top_language_stats} reports language frequency across all benchmark questions. The distribution is strongly long-tailed: a small set of languages appears frequently, while many languages appear comparatively rarely. This reflects the benchmark design goal of preserving multilingual breadth while retaining sparse-evidence conditions common in low-resource evaluation.

\begin{figure*}[htbp]
    \centering
    \includegraphics[width=\textwidth]{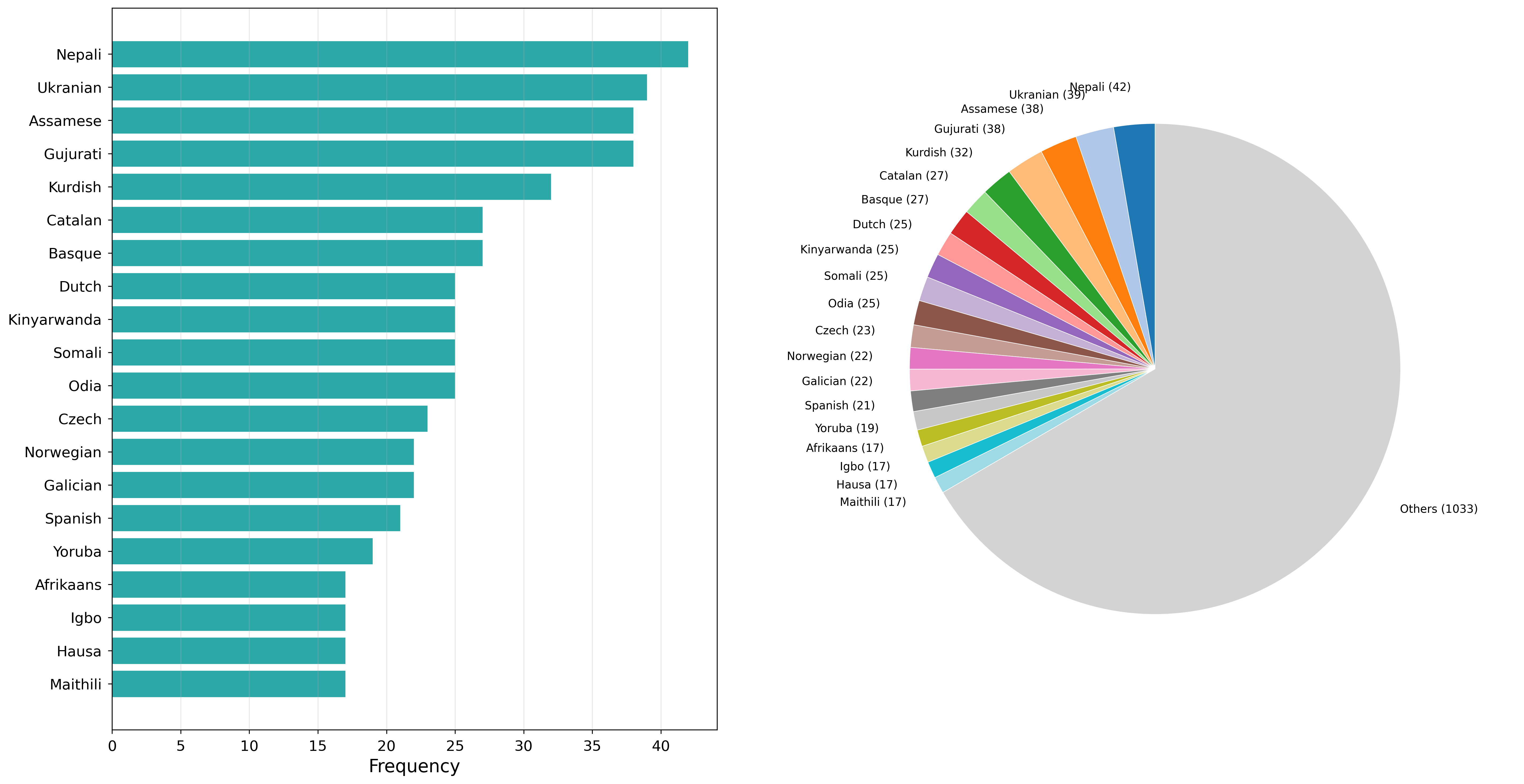}
    \caption{Language frequency across all benchmark questions. Left: top 20 languages by question frequency. Right: the same distribution with remaining languages grouped as Others.}
    \label{fig:top_language_stats}
\end{figure*}

\subsection{Model Family Distribution Across Questions}

Figure~\ref{fig:top_model_stats} reports model-family frequency across all benchmark questions. Similar to the language distribution (Figure~\ref{fig:top_language_stats}), model-family coverage is long-tailed: a few widely benchmarked families dominate, while many others appear infrequently. This imbalance mirrors real-world evaluation practice, where popular model families receive disproportionate attention, and reinforces the need for transfer-based prediction in under-evaluated settings.

\begin{figure*}[htbp]
    \centering
    \includegraphics[width=\textwidth]{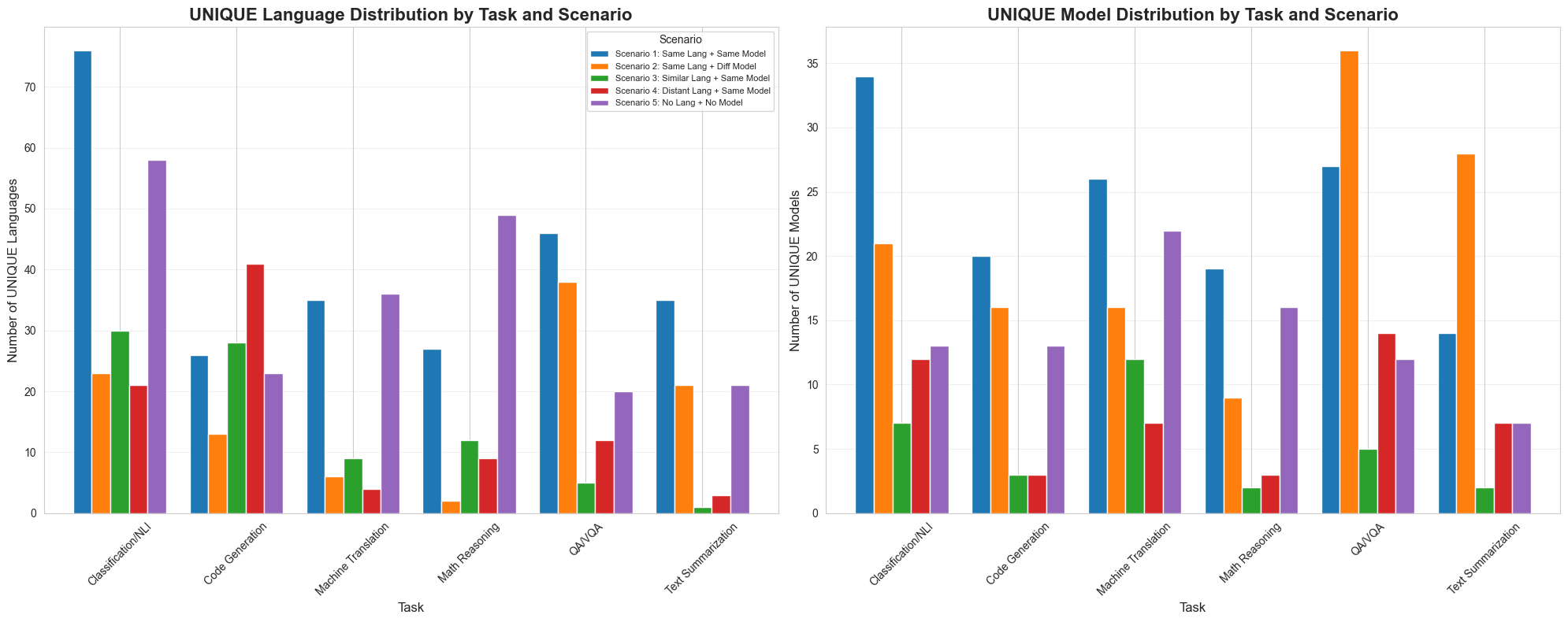}
    \caption{Model-family frequency across all benchmark questions. Left: top model families by question frequency. Right: the same distribution with remaining families grouped as Others.}
    \label{fig:top_model_stats}
\end{figure*}

\subsection{Language Similarity Computation Details}
\label{appendix:language-similarity}

We determine whether two languages are similar using language feature vectors from the \texttt{lang2vec} library with feature set \texttt{syntax\_knn+fam+geo}, which combines syntactic nearest-neighbour features, language family features, and geographic features. For languages with available vectors, we compute cosine distance between L2-normalised vectors, defined as $1 - \cos(\mathbf{u}, \mathbf{v})$.

We set the close-language threshold dynamically using the empirical distribution of pairwise distances. Specifically, we compute all pairwise cosine distances among languages with vectors and set the threshold to the 10th percentile, which yields 0.1069 in our benchmark construction. Language pairs at or below this threshold are treated as \emph{close}, and pairs above this threshold are treated as \emph{distant}. Scenario 3 selects close languages, while Scenario 4 explicitly excludes them.

\subsection{Prompt Template for Question Generation}

The prompt for question generation for \textsc{PredSet} and \textsc{QnASet} is shown below.

\begin{figure*}[htbp]
\centering
\begin{minipage}{0.98\textwidth}
\begin{boxC}
\lstset{style=promptstyle}
\begin{lstlisting}
You are given a language-to-models mapping for a specific NLP task where ground truth (GT) data is available.

Task: {task_name}
Language-Model Mappings: {mappings}

The mappings show which models have been benchmarked on which languages for this task.
Each language maps to a list of models that have been evaluated on it.

### Goal

Generate natural, varied evaluation questions for this task using the language-model combinations from the mapping.
Generate questions of the following types:

1. **Task + Model + Language**: Questions about specific model performance on specific language
2. **Task + Language**: Questions about how different models perform on a specific language
3. **Task + Model**: Questions about how a specific model performs across languages
4. **Model + Language**: Questions about model performance on language (task implied)
5. **Task only**: Questions about overall task performance across languages/models

### Question Templates (adapt and vary phrasing):

1. *What is the performance of {{model}} on {{task}} for {{language}}?*
2. *How does {{model}} perform on {{task}} in {{language}}?*
3. *Compare {{model1}} and {{model2}} performance for {{language}} on {{task}}.*
4. *Which model performs best for {{task}} in {{language}}?*
5. *What are the {{task}} results for {{model}} across all languages?*
6. *How does {{language}} performance vary across models for {{task}}?*
7. *What languages show the best performance for {{task}}?*
8. *Which models have been evaluated for {{task}} in {{language}}?*
9. *What is the cross-lingual performance of {{model}} on {{task}}?*
10. *How do low-resource languages perform on {{task}}?*

### Rules

1. **Only use languages and models from the provided mapping** - do not invent new ones. Generate questions only for natural languages. For tasks like code generation, programming languages might be present as well. DO NOT generate questions for programming languages.
2. Use the exact task name provided.
3. If multiple models/languages appear in a question, list them alphabetically, comma-separated (no spaces).
4. If task, model, or language is not present in a question, set it as `""` in the output.
5. Questions must be clear, end with `?`, and avoid hallucinations.
6. Generate diverse questions; avoid repetitive phrasing.
7. Focus on realistic evaluation scenarios where GT data would be available.

### Output Format

Return **only JSON** in this structure:

class QuestionsGenerated(BaseModel):
    class Question(BaseModel):
        complete_question: str
        task: str
        models: str
        languages: str

    questions: list[Question]

Generate {num_questions} diverse questions covering different question types and language-model combinations.
\end{lstlisting}
\end{boxC}
\end{minipage}
\caption{Prompt template used to generate illustrative questions for \textsc{PredSet} (numeric prediction) and \textsc{QnASet} (comparative reasoning).}
\label{fig:prompt_template}
\end{figure*}

\subsection{LLM-as-Judge Prompt for Quality Metrics}

The following prompt is used to evaluate system responses on four quality dimensions: predictive plausibility, feature selection, coherence, and citation emphasis. Each response report is scored on a 1--5 scale by GPT-4.1 acting as an automated judge.

\begin{figure*}[htbp]
\centering
\begin{minipage}{0.98\textwidth}
\begin{boxC}
\lstset{style=promptstyle}
\begin{lstlisting}
You are an expert reviewer: a senior multilingual-NLP researcher and applied statistician.
Your job: take a single input -- a research report -- and produce an automatic, objective evaluation along four axes:
Predictive plausibility, Feature selection, Coherence, and Citation emphasis.
For each axis return an integer score 1-5 plus a short rationale and up to 3 concrete indicators.

Scoring rubric (1-5):

1) Predictive plausibility:
   5: Clear assumptions; modeling choices justified; uncertainty quantified; sanity checks present.
   4: Good justification; minor gaps (e.g., limited uncertainty discussion).
   3: Some plausible ideas but notable missing justifications or contradictory reasoning.
   2: Weak or ad-hoc predictive reasoning; unexplained leaps.
   1: Implausible claims, contradictory or absent reasoning.

2) Feature selection:
   5: Expert-level: linguistically justified features, selection method described, interaction effects explored.
   4: Strong feature list with reasonable justifications; some missing depth.
   3: Useful features but shallow rationale; key multilingual features missing.
   2: Generic or ill-suited features; little reasoning.
   1: No coherent feature rationale; features absent or irrelevant.

3) Coherence:
   5: Logical flow; claims follow from premises; clear definitions; few language errors.
   4: Mostly coherent; minor organizational or phrasing issues.
   3: Fragmented arguments or occasional contradictions; several clarity issues.
   2: Disorganized, hard to follow, or contradictory statements.
   1: Incoherent, contradictory, or unintelligible.

4) Citation emphasis:
   5: Key claims tied to relevant citations; literature justifies design choices.
   4: Many claims cited, but a few unsupported assertions remain.
   3: Some grounding, but important claims lack citations.
   2: Sparse citation support; many claims unsupported.
   1: Virtually no citation grounding; unusual claims without references.

Output format (strict JSON only):
{
  "metrics": [
    {"metric_name": "predictive_plausibility", "score": 5,
     "rationale": "...", "indicators": ["...", "...", "..."]},
    {"metric_name": "feature_selection", "score": 4,
     "rationale": "...", "indicators": ["..."]},
    {"metric_name": "coherence", "score": 3,
     "rationale": "...", "indicators": ["..."]},
    {"metric_name": "citation_emphasis", "score": 2,
     "rationale": "...", "indicators": ["..."]}
  ],
  "overall_recommendation": {
    "average_score": 3.50,
    "verdict": "Minor revision",
    "top_actionable_improvements": ["...", "...", "..."]
  }
}

Verdict mapping: >=4.25 Accept, 3.25-4.24 Minor revision,
2.5-3.24 Major revision, <2.5 Reject.
Rationales max 50 words. Indicators max 20 words each, up to 3 per metric.
If a metric cannot be assessed, set score to null with one-line rationale.

Report: {report}
\end{lstlisting}
\end{boxC}
\end{minipage}
\caption{LLM-as-judge prompt for evaluating system responses on four quality dimensions (predictive plausibility, feature selection, coherence, citation emphasis). Each metric is scored 1--5 with structured rationale and indicators.}
\label{fig:judge_prompt}
\end{figure*}

\subsection{Prediction Extraction Prompt}

After each system generates a response report, predictions are extracted using the following LLM prompt. This prompt parses the agent's report to produce structured predictions in a format compatible with ground-truth evaluation (MAE for numeric predictions, accuracy for comparative questions).

\begin{figure*}[htbp]
\centering
\begin{minipage}{0.98\textwidth}
\begin{boxC}
\lstset{style=promptstyle}
\begin{lstlisting}
Extract agent's predicted answer from research report.

USER QUERY: {user_query}
QUERY TYPE: {predictive_or_qna}

AGENT RESEARCH REPORT:
{agent_report}

INSTRUCTIONS:

1. DETERMINE is_answer_present:
   - Set to true ONLY if agent's report contains a relevant, clear answer
   - Set to false if no answer found or report is unclear/irrelevant

2. FOR PREDICTIVE QUERIES:
   - Extract ALL performance metrics mentioned (e.g., accuracy, pass@1, BLEU, ROUGE, F1)
   - For EACH metric found, create an object with:
     * metric_name: exact name of the metric (string)
     * value: original value AS STATED in the report (string)
     * value_in_100_range: numeric value scaled to 0-100 range (float)
   - Scaling rules for value_in_100_range:
     * If value is percentage (e.g., "61.25%"): extract number -> 61.25
     * If value is decimal 0-1 (e.g., "0.85"): multiply by 100 -> 85.0
     * If value already 0-100 (e.g., "85.5"): use as-is -> 85.5

3. FOR QNA QUERIES:
   - Extract CONCISE answer value(s) from agent's report:
     * For model questions: extract model name ONLY (e.g., "GPT-4")
     * For language questions: extract language name(s) ONLY
     * For numeric questions: extract number(s) ONLY
   - NEVER include full sentences or explanations

4. BOTH PREDICTIVE AND QNA:
   - If the question can be answered BOTH ways (metrics AND concise answer):
     * Provide BOTH predicted_metrics_and_values_for_predictive AND answer_text_for_qna

OUTPUT FORMAT (strict JSON, no markdown):

For PREDICTIVE:
{
  "is_answer_present": true,
  "predicted_metrics_and_values_for_predictive": [
    {"metric_name": "pass@1", "value": "61.25", "value_in_100_range": 61.25}
  ],
  "answer_text_for_qna": ""
}

For QNA:
{
  "is_answer_present": true,
  "predicted_metrics_and_values_for_predictive": [],
  "answer_text_for_qna": "GPT-4"
}

CRITICAL RULES:
- Use ONLY information from agent's report
- value_in_100_range must be NUMERIC ONLY (float, no % or strings)
- answer_text_for_qna must be CONCISE - NO full sentences
- predicted_metrics_and_values_for_predictive is always a LIST ([] if empty)
- answer_text_for_qna is always a STRING ("" if empty)
- Output valid JSON only
\end{lstlisting}
\end{boxC}
\end{minipage}
\caption{Prompt used to extract structured predictions from system response reports. The extracted predictions are compared against ground truth to compute PredSet MAE (numeric prediction) and QnASet accuracy (comparative reasoning).}
\label{fig:prediction_extraction_prompt}
\end{figure*}

\subsection{Thought Creator Evaluation Prompt}

The following prompt evaluates whether hypotheses generated by the ThoughtCreatorAgent faithfully reflect expert guidance and stay within system capabilities. This produces the thought faithfulness and capability compliance metrics reported in Table~\ref{tab:agent_quality}.

\begin{figure*}[htbp]
\centering
\begin{minipage}{0.98\textwidth}
\begin{boxC}
\lstset{style=promptstyle}
\begin{lstlisting}
--- Faithfulness Reflection ---
You are evaluating whether thought hypotheses correctly reflect reasoning guidance.

Expert Knowledge/Guidance (from user_proxy message):
{user_proxy_message}

Generated Thoughts:
{thought_paths}

For EACH thought, determine:
- Does the thought's method/approach align with the reasoning steps suggested in the expert knowledge?
- Does it follow the strategic guidance provided?

Respond with JSON:
{"evaluations": [{"thought_name": "<name>", "is_faithful": true/false, "explanation": "<brief reason>"}]}

Be strict: mark false if the thought significantly deviates from suggested approaches or ignores key guidance.

--- System Capability Adherence ---
PROHIBITED suggestions (mark as non-compliant if suggested):
- Fine-tuning models
- Creating new datasets
- Downloading models locally
- Training new models from scratch
- Accessing external APIs not already available

For EACH thought, check the "method" field:
- Does it suggest any PROHIBITED operations?

Respond with JSON:
{"evaluations": [{"thought_name": "<name>", "is_compliant": true/false, "violations": ["violation1", ...]}]}
\end{lstlisting}
\end{boxC}
\end{minipage}
\caption{Prompts used to evaluate ThoughtCreatorAgent output: faithfulness reflection (alignment with expert guidance) and system capability adherence (avoidance of prohibited operations).}
\label{fig:thought_creator_prompt}
\end{figure*}

\subsection{Generated Code Evaluation Prompt}

The following prompt evaluates the quality of Python code generated by the Coder agent. It assesses algorithm appropriateness, feature engineering, methodology rigor, code quality, and research alignment, producing the metrics reported in the Coder Agent Analysis appendix.

\begin{figure*}[htbp]
\centering
\begin{minipage}{0.98\textwidth}
\begin{boxC}
\lstset{style=promptstyle}
\begin{lstlisting}
You are an expert code reviewer specializing in data science, ML, and research methodology.

Context:
- Task Domain: {task_context}
- User Question: {user_question}

Python Code to Analyze:
{code_content}

Assess the following dimensions:

0. TASK-QUERY ALIGNMENT: Does the code address the user's research question?
1. ALGORITHMS & MODELS: Identify all algorithms; assess appropriateness and correctness.
2. FEATURES & VARIABLES: List features; assess engineering sophistication and validity.
3. METHODOLOGY: Identify approach; assess statistical rigor and evaluation strategy.
4. CODE QUALITY: Assess organization, error handling, and best practices.
5. RESEARCH APPROPRIATENESS: Is the methodology scientifically sound?
6. SOPHISTICATION: Rate as Basic, Intermediate, or Advanced.

Output JSON:
{
  "algorithms_used": ["..."], "algorithm_appropriateness": "appropriate|questionable|inappropriate",
  "features_used": ["..."], "feature_engineering_level": "none|basic|moderate|advanced",
  "methodology_type": "regression|classification|...", "methodology_rigor": "high|moderate|low",
  "code_quality": "excellent|good|fair|poor",
  "task_query_alignment": "well_aligned|moderately_aligned|poorly_aligned",
  "sophistication_level": "basic|intermediate|advanced",
  "overall_assessment": "excellent|good|fair|poor",
  "strengths": ["..."], "weaknesses": ["..."], "summary": "1-2 sentences"
}
\end{lstlisting}
\end{boxC}
\end{minipage}
\caption{Prompt used to evaluate generated Python code quality across seven dimensions: task alignment, algorithms, features, methodology, code quality, research soundness, and sophistication.}
\label{fig:code_eval_prompt}
\end{figure*}

\subsection{Tool Call Relevance Evaluation Prompt}

The following prompt evaluates whether web search and crawl tool calls are relevant to the active hypothesis. This produces the web search relevance metric reported in Table~\ref{tab:agent_quality}.

\begin{figure*}[htbp]
\centering
\begin{minipage}{0.98\textwidth}
\begin{boxC}
\lstset{style=promptstyle}
\begin{lstlisting}
You are evaluating whether a web search/crawl tool call query is relevant
for answering a specific research thought/hypothesis.

Thought Name: {thought_name}

Thought Context:
Hypothesis: {hypothesis}
Method: {method}
Background: {background}

Tool Call Query: {tool_query}

A query is RELEVANT if:
- It directly searches for information needed to test the hypothesis
- It seeks data, benchmarks, or evidence mentioned in the thought's method
- It looks for papers, documentation, or results that would help answer the thought
- It searches for related concepts, models, or comparisons needed for analysis

A query is NOT RELEVANT if:
- It searches for completely unrelated topics
- It duplicates information already available in the thought context
- It's too vague or generic to be useful
- It searches for information not needed for this specific thought

Return JSON:
{
  "query": "{tool_query}",
  "thought_name": "{thought_name}",
  "is_relevant": true/false,
  "reasoning": "Brief explanation (1-2 sentences)"
}
\end{lstlisting}
\end{boxC}
\end{minipage}
\caption{Prompt used to evaluate web search and crawl tool call relevance against the active hypothesis context.}
\label{fig:tool_relevance_prompt}
\end{figure*}

\section{Use of AI Assistants}

AI writing assistants were used during the preparation of this manuscript for paraphrasing and grammar correction. All scientific content, experimental design, analysis, and conclusions are the sole work of the authors.

\end{document}